\documentclass{article}

\usepackage{microtype}
\usepackage{graphicx}
\usepackage{subfigure}
\usepackage{booktabs} 

\usepackage{hyperref}

\usepackage[accepted]{icml2024}

\usepackage{amsmath}
\usepackage{amssymb}
\usepackage{mathtools}
\usepackage{amsthm}

\usepackage[capitalize,noabbrev]{cleveref}

\theoremstyle{plain}

\theoremstyle{definition}

\theoremstyle{remark}

\usepackage{booktabs}
\usepackage{multirow}
\usepackage[normalem]{ulem}
\useunder{\uline}{\ul}{}
\usepackage{subfigure}
\usepackage{tcolorbox} 
\usepackage{listings}  

\usepackage[textsize=tiny]{todonotes}

\icmltitlerunning{DS-Agent: Automated Data Science by Empowering Large Language Models with Case-Based Reasoning}

\begin{document}

\twocolumn[
\icmltitle{DS-Agent: Automated Data Science\\ by Empowering Large Language Models with Case-Based Reasoning}




\icmlsetsymbol{equal}{*}

\begin{icmlauthorlist}
\icmlauthor{Siyuan Guo}{SAI,ERC,ICFS}
\icmlauthor{Cheng Deng}{SJTU}
\icmlauthor{Ying Wen}{SJTU}
\icmlauthor{Hechang Chen}{SAI,ERC}
\icmlauthor{Yi Chang}{SAI,ERC,ICFS}
\icmlauthor{Jun Wang}{UCL}

\end{icmlauthorlist}

\icmlaffiliation{SAI}{School of Artificial Intelligence, Jilin University}
\icmlaffiliation{ERC}{Engineering Research Center of Knowledge-Driven Human-Machine Intelligence, Jilin University}
\icmlaffiliation{ICFS}{International Center of Future Science, Jilin University}
\icmlaffiliation{SJTU}{Shanghai Jiao Tong University}
\icmlaffiliation{UCL}{University College London}

\icmlcorrespondingauthor{Hechang Chen}{chenhc@jlu.edu.cn}
\icmlcorrespondingauthor{Yi Chang}{yichang@jlu.edu.cn}
\icmlcorrespondingauthor{Jun Wang}{jun.wang@cs.ucl.ac.uk}

\icmlkeywords{Machine Learning, ICML}

\vskip 0.3in
]



\printAffiliationsAndNotice{}  

\begin{abstract}
In this work, we investigate the potential of large language models (LLMs) based agents to automate data science tasks, with the goal of comprehending task requirements, then building and training the best-fit machine learning models.
Despite their widespread success, existing LLM agents are hindered by generating unreasonable experiment plans within this scenario.
To this end, we present DS-Agent, a novel automatic framework that harnesses LLM agent and case-based reasoning (CBR). 
In the development stage, DS-Agent follows the CBR framework to structure an automatic iteration pipeline, which can flexibly capitalize on the expert knowledge from Kaggle, and facilitate consistent performance improvement through the feedback mechanism. 
Moreover, DS-Agent implements a low-resource deployment stage with a simplified CBR paradigm to adapt past successful solutions from the development stage for direct code generation, significantly reducing the demand on foundational capabilities of LLMs. 
Empirically, DS-Agent with GPT-4 achieves 100\% success rate in the development stage, while attaining 36\% improvement on average one pass rate across alternative LLMs in the deployment stage. In both stages, DS-Agent achieves the best rank in performance, costing \$1.60 and \$0.13 per run with GPT-4, respectively.
Our data and code are open-sourced at \url{https://github.com/guosyjlu/DS-Agent}.
\end{abstract}
\section{Introduction}
Recently, the remarkable foundational capabilities of large language models (LLMs) \cite{chatgpt, gpt-4} have empowered autonomous language agents to address a wide spectrum of tasks effectively \cite{saycan,computer-task,hugginggpt,chemical-research,mathematical-discoveriy}. In this work, we explore an open-ended decision making scenario-automated data science \cite{automating-ds, chatgpt-ds}, which aims at democratizing access to data insights while minimizing the need for specialized expertise. Specifically, our focus is on automating machine learning (ML), a particularly specialized part, with the primary goal of comprehending task requirements, building and training the best-fit ML models, and finally deploying the trained model. 

Despite the widespread success of LLM agents, a recent work \cite{mlagentbench} indicates that existing agents, including AutoGPT \cite{autogpt}, LangChain \cite{langchain}, and the state-of-the-art ResearchAgent \cite{mlagentbench}, struggle to achieve a high task completion rate within the data science scenario, even when implemented with the most powerful LLM GPT-4. This is mainly attributed to LLMs' deficiency in generating reasonable plans and their hallucination issues. To mitigate this, a promising solution is to further finetune LLMs to align with the automated data science scenario \cite{glam,agent-tuning,fireact,pangu}. Nevertheless, collecting sufficient samples for finetuning poses a significant challenge due to the time costs involved, particularly because feedback from automated data science tasks necessitates the completion of code execution. Worse still, since LLMs typically have billions of parameters, the back-propagation and optimization during the finetuning leads to intensive computation resources.

In this context, Kaggle emerges as a pivotal resource. As the world's largest data science competition platform, it boasts a vast repository of technical reports and codes contributed by a community of seasoned data scientists. To empower LLM agents to harness this wealth of expert knowledge efficiently, we turn to a classical AI problem-solving paradigm--case-based reasoning (CBR) \cite{cbr-review-1, cbr-review-2}. The CBR framework operates by retrieving similar past problems, reusing their solutions for the current problem, evaluating the effectiveness, revising the solution, and retaining successful solutions. Utilizing CBR enables LLM agents to analyze, extract and reuse solution patterns from these human insights, and to iteratively revise the solution based on the execution feedback to attain consistently improved performance. The integration of CBR into LLM agents not only enhances their problem-solving abilities in data science tasks but also achieves high efficiency in both sample and computation resources.

\begin{figure}[t]
  \centering
  \subfigure[]{\includegraphics[width=0.53\linewidth]{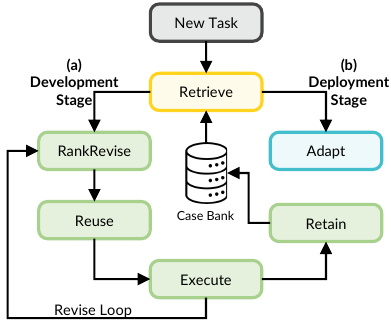}}
  \subfigure[]{\includegraphics[width=0.455\linewidth]{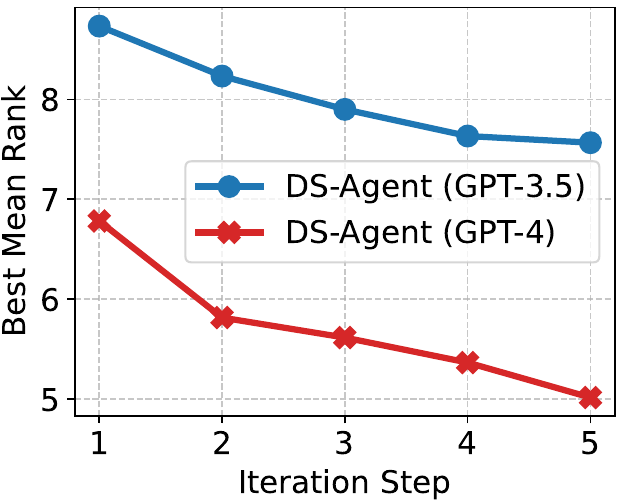}}
  \caption{(a) Overview of DS-Agent with CBR based LLMs. (b) Performance improvement of DS-Agent with increasing iteration steps by CBR over 12 development tasks.}
  \label{fig:merge-overview-exploration}
\end{figure}

To this end, we propose DS-Agent, a novel framework that harnesses LLM agent and CBR to facilitate model-centric automated data science, as depicted in \autoref{fig:merge-overview-exploration} (a).
Overall, DS-Agent operates in two distinct stages: the standard development stage, and low-resource deployment stage. 
For the development stage, DS-Agent builds on top of CBR framework to capitalize on the collected human insights from Kaggle, structuring an automatic iteration pipeline. Given a new task, DS-Agent retrieves and reuses relevant human insights from Kaggle to develop an experiment plan, and then iteratively adjusts the retrieved case and revises the experiment plan in response to the execution feedback. Benefiting from the CBR framework, DS-Agent can leverage expert knowledge from Kaggle to develop grounded experiment plans, while providing a flexible learning mechanism by retaining successful solutions to the case bank instead of resource-intensive parameter updates through back-propagation. Moreover, the feedback mechanism in CBR allows DS-Agent to iteratively retrieve useful cases and revise the experiment plan, achieving consistent performance improvement, as shown in \autoref{fig:merge-overview-exploration} (b).

In the deployment stage, DS-Agent employs a simplified CBR framework for a low-resource scenario, where the task is code generation directly in response to the user's task requirements without iterative revise based on execution feedback. Particularly, DS-Agent retrieves and reuses past successful solutions collected from the development stage for the current task. As such, DS-Agent benefits from the simplified CBR framework to facilitate the knowledge transfer from past solutions to solve an unseen deployment task in the same task distribution. With a similar solution case in the context, DS-Agent necessitates only making minor modifications for adaptation, thereby significantly reducing the demands on foundational capabilities of LLMs.

Empirically, we demonstrate the superiority of DS-Agent across \emph{30} data science tasks in both stages. During the development stage, DS-Agent with GPT-4 achieves 100\% success rate across \emph{12} tasks. For the deployment stage, DS-Agent with GPT-3.5 and GPT-4 achieves 85\% and 99\% one pass rate over \emph{18} deployment tasks, while the best baseline attains a mere 56\% and 60\%. Remarkably, DS-Agent improves the one pass rate of an open-source LLM \texttt{Mixtral-8x7b-Instruct} from a mere 6\% to 31\%. In both stages, DS-Agent with GPT-4 and GPT-3.5 attains the highest and second-highest ranks in performance. Moreover, DS-Agent costs \$0.06 and \$1.60 per run for GPT-3.5 and GPT-4 with standard scenario, which reduce further to \$0.0045 and \$0.135 in low-resource scenarios, rendering DS-Agent highly appealing for real-world deployment.

\section{Preliminary}

\begin{figure}[t]
\begin{center}    \centerline{\includegraphics[width=0.9\linewidth]{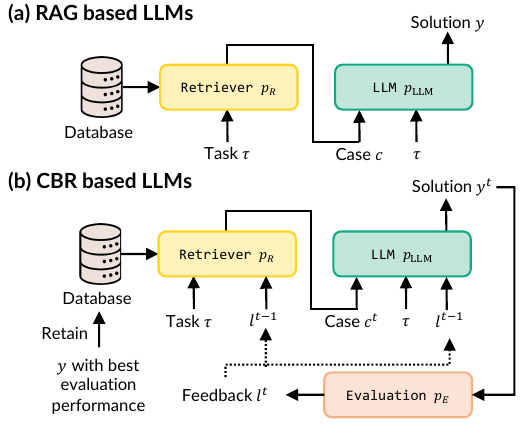}}
\caption{Comparison between (a) RAG based LLMs and (b) CBR based LLMs.}
\label{fig:rag-cbr}
\end{center}
\end{figure}

\noindent\textbf{CBR based LLMs.} 
CBR \cite{cbr-review-1, cbr-review-2} is a classical AI paradigm, which solves a new task by retrieving other similar problems, reusing their solutions, evaluating the effectiveness, and iteratively revising the solution as needed. The solution with best evaluation performance is retained to the database for future reuse. In this work, we integrate the CBR framework into LLMs to enhance their problem-solving capabilities. As shown in \autoref{fig:rag-cbr} (b), CBR based LLMs encompass three components: (i) a retriever $p_{\text{R}}$ that returns the distributions over the database based on the task $\tau$ and feedback $l$, (ii) an LLM $p_{\text{LLM}}$ that generates the solution $y$ with the task $\tau$, feedback $l$ and the retrieved case $c$, and (iii) an evaluator $p_{\text{E}}$ that produces feedback $l$ of the solution $y$. Formally, CBR based LLMs encompass an iteration loop with the $t$-th step as

\vspace{-2em}
{\small
\begin{equation}
\label{cbr}
\begin{aligned}
    & p_{\text{CBR}}(y^t|\tau) = \\
    &\sum_{l^{t-1}} p_{\text{E}}(l^{t-1}|\tau)
    \sum_{c^t}{p_{\text{R}}(c^t|\tau, l^{t-1})p_{\text{LLM}}(y^t|c^t,\tau,l^{t-1})},
\end{aligned}
\end{equation}
}
where the solution distribution is marginalized by execution feedback of the last step $l^{t-1}$ and the retrieved case $c^t$. Then, the feedback distribution can be formulated as
\begin{equation}
\begin{aligned}
    p_{\text{E}}(l^{t}|\tau) = \sum_{y^{t}}p_{\text{CBR}}(y^{t}|\tau)p_{\text{E}}(l^{t}|y^{t},\tau).
\end{aligned}
\end{equation}
As such, the feedback is produced by evaluating the current solution distribution, and the subsequent solution distribution is revised based on the current feedback, thus forming an iteration loop with consistent performance improvement.

\noindent\textbf{Comparison with Retrieval-Augmented Generation.} 
CBR based LLMs exhibit similarity with retrieval-augmented generation (RAG) \cite{rag-1, rag-2, rag-survey} - both involve the retrieval and reuse. As shown in \autoref{fig:rag-cbr} (a), RAG based LLMs only involve a retriever and an LLM, and can be formulated as
\begin{equation}
\label{rag}
    p_{\text{RAG}}(y|\tau) = \sum_{c}{p_{\text{R}}(c|\tau)p_{\text{LLM}}(y|c,\tau)},
\end{equation}
where the solution distribution is marginalized by a single latent variable, i.e., the retrieved case $c$. Therefore, while both LLMs can retrieve and reuse solution patterns from the retrieved case, CBR based LLMs can additionally adjust the retrieved the case and revise the solution in response to the evaluation feedback. Moreover, retaining good solutions to the database enables CBR based LLMs to achieve a flexible learning mechanism, thereby leading to consistent performance improvement.

\begin{figure*}[t]
    \begin{center}
    \centerline{\includegraphics[width=\textwidth]{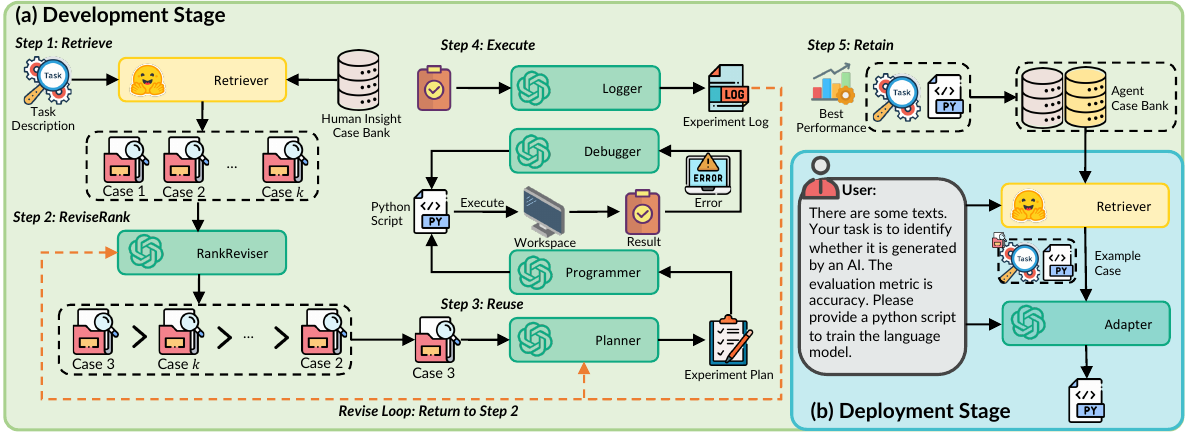}}
    \caption{The diagram of DS-Agent. \textbf{(a) Development Stage:} DS-Agent structures an automatic iteration pipeline to build and revise the model based on execution feedback. \textbf{(b) Deployment Stage:} DS-Agent adapts past successful solutions for code generation.}
    \label{fig:overview}
    \end{center}
\end{figure*}

\section{The DS-Agent}
In this section, we present DS-Agent, an automatic framework that leverages LLM agent and CBR to solve data science tasks. As shown in \autoref{fig:overview}, DS-Agent operates in two stages: the development stage and the deployment stage. Each stage handles the corresponding task set $\mathcal{T}_{\text{develop}}$ and $\mathcal{T}_{\text{deploy}}$, where the task is defined as a tuple $\left(\tau, \mathcal{D}_{\text{train}}, \mathcal{D}_{\text{valid}}, \mathcal{D}_{\text{test}}, \mathcal{M}\right)$. For both stages, DS-Agent comprehends the task description $\tau$, generates code to train an ML model with the training set $\mathcal{D}_{\text{train}}$, and evaluates its performance on the validation set $\mathcal{D}_{\text{valid}}$ using the evaluation metric $\mathcal{M}$. We report the performance of the trained ML models on the test set $\mathcal{D}_{\text{test}}$. 

\subsection{Development Stage: Automatic Iteration Pipeline}
\label{sec:stage-I}
In the development stage, we structure the workflow of DS-Agent to emulate the iterative process a data scientist follows of building, training, and validating ML models given a data science task. However, since LLMs are not inherently trained for data science scenarios, they lack precise knowledge to generate reasonable plans for the design of ML models, which leads to an unreliable task completion rate \cite{mlagentbench}. Kaggle, as a leading platform for data science competitions, offers a rich repository of expert insights and solutions that employ cutting-edge ML techniques. By integrating these practical, expert examples into LLM agents, we can significantly improve their capability to solve complex data science tasks. To this end, we propose to integrate CBR into the automatic iteration pipeline of DS-Agent as shown in \autoref{fig:overview} (a). Now, we elaborate on the automatic iteration pipeline of DS-Agent as follows.

\noindent\textbf{Human Insight Case Collection.}
Our primary objective is to collect expert insights and solutions with advanced ML techniques from Kaggle. Concretely, we select several recently completed Kaggle competitions, concentrating on three data modalities: text, time series and tabular data. This aligns with the modalities present in the development and deployment task set in this work. From selected competitions, we crawl both the technical reports shared by the winning teams and the codes with top-ranked scores in the public leaderboard. These materials undergo a reformulation process: technical reports are cleaned to reserve core insights, while the code is summarized using GPT-3.5 to convert complex implementation into textual insights. Then, they are stored into the human insight case bank.

\noindent\textbf{Step 1: Retrieve.}
First, DS-Agent retrieves relevant cases from the human insight case bank $\mathcal{C}$ that pertain to the current data science task. Particularly, \texttt{Retriever} calculates the similarity between the task description $\tau$ and case $c \in \mathcal{C}$ with their cosine similarity: $\text{sim}(\tau,c) = \cos(\mathbf{E}(\tau), \mathbf{E}(c))$, where $\mathbf{E}(\cdot)$ denotes the pretrained embedding model. Then, the top-$k$ cases with the highest similarities to the task description are retrieved in this step.

\noindent\textbf{Step 2: ReviseRank.}
While the above step generally ensures the relevance of the retrieved cases, it cannot dynamically adjust the retrieved cases in response to the execution feedback of the previous iteration. One possible solution is to finetune the retriever using the execution feedback \cite{replug, llm-r}. Yet, automated data science tasks present a unique challenge as they require code execution to produce feedback, incurring considerable time and computational expenses. To address this issue, we propose to harness the LLMs' capability to estimate the utility of the retrieved cases by analyzing the execution feedback, and then revise the ranking order to adjust the retrieved case.

Inspired by a recent work \cite{llm-rerank} that leverages LLMs for relevance ranking in web search scenarios, we adopt a similar prompt format to assign each of top-$k$ retrieved cases $\{c_1, c_2, ..., c_k\}$ with a unique identifier (e.g., [1], [2], etc.). We then prompt LLMs to generate the permutation of theses cases in descending order of their estimated utility for the current data science task, as informed by the feedback from the last iteration. The ranking results are generated in the format of [2] \textgreater [1] \textgreater [3], etc. Formally, at the iteration step $t$, the utility distribution of each case is estimated as $p_{\text{RR}}(c|\tau, l^{t-1})=p_{\text{LLM}}(c|c_1, c_2, ..., c_k, \tau, l^{t-1})$, where $p_{\text{LLM}}$ denotes the distribution of the LLM, $l^{t-1}$ denotes the execution feedback in the iteration step $t-1$, initialized as an empty string, i.e., $l^0=\emptyset$. Then, the top-ranked case $c^{t}$ is proceeded for the next step of reuse. Consequently, within the automatic iteration pipeline, the case for reuse is iteratively refined by \texttt{ReviseRanker} in response to the execution feedback. This dynamic adjustment provides DS-Agent with iteratively updated foundational materials to revise the solution for the experimental plan.

\noindent\textbf{Step 3: Reuse.}
In this phase, DS-Agent employs \texttt{Planner} to reuse the retrieved case to develop the solution for the experiment plan. In the iteration step $t$, \texttt{Planner} examines the task description $\tau$ and the previous execution feedback $l^{t-1}$ to comprehend the current context. Then, it thoughtfully analyzes the top-ranked case $c^t$, reuses the human insights included to adapt to the current task, and finally develops a new solution of the experiment plan $y^t$.

\noindent\textbf{Step 4: Execute.} 
Subsequently, DS-Agent implements the experiment plan with a Python script, and executes it to derive the empirical feedback. In particular, \texttt{Programmer} goes through the task description and experiment plan to generate the corresponding Python code. Following code generation, the script is executed to review the output. If errors are reported, \texttt{Debugger} is employed to identify and resolve bugs. Taking inspirations from Reflexion \cite{reflexion}, \texttt{Debugger} first reflects on potential bugs according to the execution feedback, and then generates and re-execute the corrected code. This debugging process continues until there are no further errors reported, or until the maximum number of predefined debugging attempts has been exhausted. Finally, \texttt{Logger} outputs a comprehensive summary of the concluded experiment's progress and results of natural language form. The preservation of the experiment log provides DS-Agent with the execution feedback, enabling it to further revise the experiment plan to design better ML models for the current task.

\noindent\textbf{Step 5: Retain.}
At the end of each iteration step, we utilize the trained ML model to make predictions on the test set. If improved performance is observed, DS-Agent archives the task description $\tau$ along with the corresponding Python script $s$ into both human insight case bank $\mathcal{C}$ and agent case bank $\mathcal{B}$ as an example solution case for future reuse.

\noindent\textbf{Revise Loop: Return to Step 2.}
After the Retain step, the workflow returns to the ReviseRank step to create a Revise loop. This loop empowers DS-Agent to further revise the solution for the experiment plan according to the execution feedback of the current step $l^t$. The Revise loop terminates when the predetermined maximum number of iteration steps has been reached.

With the aforementioned steps during the development stage, DS-Agent utilizes the CBR framework to iteratively retrieve and reuse relevant, effective case to revise the solution of the experiment plan, leading to improved problem-solving capabilities for data science tasks. Here, the CBR framework can be formulated as

\vspace{-0.5em}
{\small
\begin{equation}
\label{cbr_dev}
\begin{aligned}
    & p_{\text{CBR}}^{\text{dev}}(y^t|\tau) = 
    \sum_{l^{t-1}}p_{\text{E}}(l^{t-1}|\tau) \cdot \\ &\sum_{c \in \text{top-}k(\text{sim}(\tau, \cdot))} 
    p_{\text{RR}}(c \mid \tau, l^{t-1}) p_{\text{LLM}}(y^t \mid \tau, c, l^{t-1}),  
\end{aligned}
\end{equation}
}
which aligns with the solution distribution of CBR based LLMs in \autoref{cbr} and the only difference is that we utilize both \texttt{Retriever} and \texttt{ReviseRanker} to retrieve cases in response to the task and the execution feedback.

We summarize the pseudo-code of the automatic pipeline in Algorithm \ref{alg:development}. Overall, DS-Agent benefits from the CBR paradigm in two aspects. Firstly, CBR integrates the human insight case bank, which contains intensive expert knowledge of the data science, enabling DS-Agent to derive reasonable experiment plans. Moreover, CBR offers a flexible learning mechanism by retaining successful solution cases into the human insight case bank, thus eliminating the need for resource-intensive finetuning of LLMs via back-propagation. For instance, when encountering novel tasks that involve previously unseen data modalities, such as graph data \cite{pei2020active, pei2024hago, pei2024memory}, it can simply integrate the latest human insights into the case bank $\mathcal{C}$. This enables DS-Agent to adeptly solve data science tasks related to graph data by drawing on its enriched knowledge repository.

Secondly, the Revise loop within CBR allows DS-Agent to utilize the execution feedback from the last iteration to guide the case retrieval and to revise the experiment plan via case reuse. This iterative loop leads to consistent performance improvement by progressively revising the design of ML models towards an optimal fit. We plot the performance curve of DS-Agent with the increasing iteration steps in Figure \ref{fig:merge-overview-exploration} (b). A consistent trend of performance improvement is observed empirically.

\subsection{Deployment Stage: Learning from Past Cases}
\label{sec:stage-II}
In the deployment stage, we aim to reuse the past successful solution cases archived in the agent case bank $\mathcal{B}$ to achieve a low-resource scenario, where DS-Agent directly generates the Python code in response to the user's task requirements for training ML models. Without the iteration loop, we simplify the CBR paradigm of the development stage to implement DS-Agent by adapting solution code from similar tasks to the current ones. 

As shown in Figure \ref{fig:overview}, DS-Agent first retrieves relevant case and then reuses the case to adapt to the deployment tasks. Specifically, given a deployment task $\tau$, DS-Agent first retrieves a case pair $(\tau_0, s_0)$ from the agent case bank $\mathcal{B}$ with similar task description, i.e., $(\tau_0, s_0) = \arg\max_{(\tau_0, s_0)\in\mathcal{B}} \text{sim}(\tau, \tau_0)$. Then, DS-Agent utilizes \texttt{Adapter} to reuse the retrieved example case pair for adaptation to the current task, generating the solution code for training ML models. This simplified CBR framework can be formulated as
\begin{equation}
\label{cbr_dep}
\begin{aligned}
    & p_{\text{CBR}}^{\text{dep}}(s|\tau) =  &p_{\text{LLM}}(s | \arg\max_{(\tau_0, s_0) \in \mathcal{B}} \text{sim}(\tau, \tau_0), \tau). 
\end{aligned}
\end{equation}
In the deployment stage, DS-Agent leverages a simplified CBR paradigm to facilitate the transfer of knowledge from the past successful cases to solve an unseen data science task in the same task distribution. By providing a similar solution case in the context, DS-Agent necessitates only minor modifications to tailor it to the new task. This significantly eases the demands on the reasoning and coding capabilities of LLMs. As a result, DS-Agent can be implemented on top of even open-source LLMs for the deployment stage.
\section{Experiments}
\subsection{Experiment Setting}
\noindent\textbf{Task Selection.} We select 30 data science tasks with three data modalities, including text, time series and tabular data, and two fundamental task types of regression and classification. These diverse datasets were sourced from a variety of platforms. We incorporate various evaluation metrics for these tasks. Out of the 30 tasks, 12 have been earmarked for the development stage, while the remaining 18 are designated for deployment. For each dataset, we write natural language task description, and split them into training set, validation set and the test set. Besides, we prepare a Python script that establishes a baseline of random guess, serving as an initial reference point. The detailed dataset description is presented in Table \ref{tab:task-description}.

\noindent\textbf{Evaluation Metric.} We mainly evaluate the agent's ability from three aspects:
\textbf{(1) Completion of building ML models.} In the development stage, we employ the success rate, i.e., whether the agent can build an ML model in a bug-free manner within a fixed number of steps. In the deployment stage, the metric is the one-pass rate, indicating the agent's ability to build an ML model with only a single trial.
\textbf{(2) Performance of built ML models.} For both stages, we utilize mean rank and best rank as the evaluation metric to evaluate the agents' capabilities for automated data science.
\textbf{(3) Resource cost.} Since we mainly utilize closed-source LLMs in this work, we take the consumed money to assess resource costs. 

Please refer to Appendix \ref{app:exp-details} for more experiment details.

\begin{figure}
    \begin{center}
    \centerline{\includegraphics[width=0.85\linewidth]{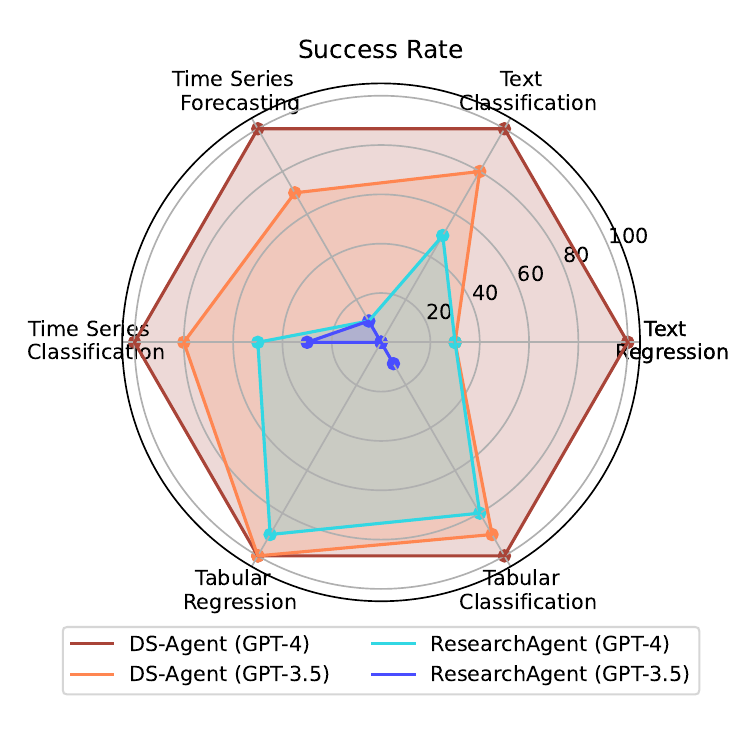}}
    \caption{Success rate of four different agents in the development stage. The reported results are averaged across five repetitive trials.}
    \label{fig:success_rate_stage_I}
    \end{center}
    \vspace{-25pt}
\end{figure}

\begin{table*}[t]
\caption{Mean rank and best rank w.r.t. task-specific evaluation metric results on 12 data science tasks in the development stage. Results are reported over five repetitive trials. Best performances are highlighted in bold, and second best performances are underlined.}
\resizebox{\linewidth}{!}{
\begin{tabular}{@{}c|l|lcccccccccccc|c@{}}
\toprule \midrule
 &  &  & \textbf{FB} & \textbf{AR} & \textbf{TE} & \textbf{CP} & \textbf{ETT} & \textbf{ILI} & \textbf{HW} & \textbf{EC} & \textbf{MCS} & \textbf{WBY} & \textbf{ST} & \textbf{ES} & \textbf{Avg} \\ \midrule
\multirow{4}{*}{\textbf{Mean Rank}} & \multirow{2}{*}{\textbf{GPT-3.5}} & \textbf{ResearchAgent} & 8.0 & 10.0 & 12.0 & 13.0 & 9.4 & 11.0 & 14.2 & 12.2 & 15.0 & 16.0 & 15.8 & 14.0 & 12.6 \\
 &  & \textbf{DS-Agent} & {\ul 7.4} & {\ul 8.2} & {\ul 6.2} & {\ul 7.2} & {\ul 7.2} & {\ul 8.2} & {\ul 6.4} & 10.2 & \textbf{6.2} & {\ul 6.0} & {\ul 7.4} & 9.6 & {\ul 7.5} \\ \cmidrule(l){2-16} 
 & \multirow{2}{*}{\textbf{GPT-4}} & \textbf{ResearchAgent} & 7.6 & 8.6 & 10.6 & 11.8 & 10.0 & 9.4 & 12.6 & {\ul 7.2} & 10.4 & 10.0 & 10.6 & {\ul 9.2} & 9.8 \\
 &  & \textbf{DS-Agent} & \textbf{3.4} & \textbf{4.2} & \textbf{5.8} & \textbf{4.4} & \textbf{4.4} & \textbf{4.4} & \textbf{5.4} & \textbf{6.6} & {\ul 6.8} & \textbf{5.6} & \textbf{4.4} & \textbf{4.4} & \textbf{5.0} \\ \midrule \midrule
\multirow{4}{*}{\textbf{Best Rank}} & \multirow{2}{*}{\textbf{GPT-3.5}} & \textbf{ResearchAgent} & 8.0 & 10.0 & 12.0 & 13.0 & 7.0 & 11.0 & 12.0 & 9.0 & 15.0 & 16.0 & 15.0 & 14.0 & 11.8 \\
 &  & \textbf{DS-Agent} & {\ul 5.0} & {\ul 2.0} & {\ul 2.0} & {\ul 3.0} & {\ul 3.0} & 6.0 & \textbf{1.0} & 7.0 & {\ul 2.0} & \textbf{1.0} & {\ul 2.0} & 6.0 & {\ul 3.3} \\ \cmidrule(l){2-16} 
 & \multirow{2}{*}{\textbf{GPT-4}} & \textbf{ResearchAgent} & 6.0 & 5.0 & 7.0 & 10.0 & 10.0 & {\ul 3.0} & 9.0 & {\ul 2.0} & \textbf{1.0} & {\ul 2.0} & 7.0 & {\ul 3.0} & 5.4 \\
 & & \textbf{DS-Agent} & \textbf{1.0} & \textbf{1.0} & \textbf{1.0} & \textbf{1.0} & \textbf{1.0} & \textbf{1.0} & {\ul 3.0} & \textbf{1.0} & 4.0 & {\ul 2.0} & \textbf{1.0} & \textbf{1.0} & \textbf{1.5} \\ \midrule \bottomrule
\end{tabular}
}
\label{tab:main-result-stage-1}
\vspace{-5pt}
\end{table*}

\subsection{Results for Development Stage}
\subsubsection{Main Results}
\noindent\textbf{Baselines.} In the development stage, we conduct a comparative analysis between DS-Agent and ResearchAgent \cite{mlagentbench}, which is the state-of-the-art language agent for solving ML research related tasks. Both agents are implemented on top of GPT-3.5 and GPT-4, respectively.

\noindent\textbf{Comparison on success rate.} 
First, we analyze the success rate of different agents in terms of six different types of data science tasks in the development stage. As shown in Figure \ref{fig:success_rate_stage_I}, DS-Agent with GPT-4 achieves the highest success rate of 100\% over all the tasks. Notably, DS-Agent with GPT-3.5 consistently surpasses ResearchAgent with GPT-4 in all tasks, underscoring the effectiveness of the proposed agent framework. Among them, ResearchAgent with GPT-3.5 almost fails in every type of task, which can be attributed to its demanding requirements for the reasoning and coding abilities of LLMs. Interestingly, agents exhibit a higher level of proficiency in tabular tasks compared to other types of tasks. This inclination can be explained by the observation that tabular tasks usually entail simply calling functions from sklearn \cite{sklearn}, demanding considerably less reasoning and coding ability from LLM agents compared to other tasks.

\noindent\textbf{Comparison on task-specific evaluation metric.} 
Next, we turn our attention to a detailed comparison based on task-specific evaluation metrics across 12 development tasks. The corresponding results are presented in Table \ref{tab:main-result-stage-1}. From the table, we can observe that DS-Agent with GPT-4 significantly outperforms other agents in terms of both mean rank and best rank. Particularly, DS-Agent with GPT-4 achieves the best performance in 9 out of 12 data science tasks. Furthermore, DS-Agent with GPT-3.5 achieves the second best average result in terms of both mean and best rank, even surpassing ResearchAgent with GPT-4 across the majority of tasks. These findings highlight the superiority of DS-Agent in solving data science tasks.

A crucial aspect of DS-Agent's design lies in the automatic iteration pipeline supported by CBR, allowing it to consistently revise the experiment plan by incorporating real feedback from code execution. To illustrate this process, we depict the average best mean rank of DS-Agent across all tasks as iteration steps increase in Figure \ref{fig:merge-overview-exploration} (b). The noticeable performance improvement of DS-Agent with both GPT-3.5 and GPT-4 over increasing iterations demonstrates the efficacy of the proposed automatic iteration pipeline.

\begin{table}[t]
\caption{Ablation results in terms of average best rank over 12 development tasks. Results are reported over five repetitive trials.}
\centering
\resizebox{0.7\linewidth}{!}{
\begin{tabular}{@{}lc@{}}
\toprule
GPT-4 & Average Best Rank \\ \midrule
DS-Agent & \textbf{2.08} \\
DS-Agent w/o ReviseRank & 2.58 \\
DS-Agent w/o CBR & 3.41 \\ \bottomrule
\end{tabular}
}
\label{tab:ablation-stage-1}
\end{table}

\begin{table*}[tbp]
\caption{Mean rank w.r.t. task-specific evaluation metric results on 18 data science tasks in the deployment stage. Results are reported over 10 repetitive runs. Best performances are highlighted in bold, and second best performances are underlined.}
\resizebox{\linewidth}{!}{
\begin{tabular}{@{}l|l|cccccccccccccccccc|c@{}}
\toprule
 &  & \textbf{JS} & \textbf{HR} & \textbf{BPP} & \textbf{WR} & \textbf{DAG} & \textbf{BQ} & \textbf{TFC} & \textbf{WTH} & \textbf{ELE} & \textbf{SRC} & \textbf{UGL} & \textbf{HB} & \textbf{CA} & \textbf{CS} & \textbf{MH} & \textbf{SS} & \textbf{CO} & \textbf{SD} & \textbf{Avg} \\ \midrule
\textbf{Mixtral} & \textbf{Zero-shot} & 37.0 & 35.0 & 35.0 & 31.0 & 35.0 & 32.0 & 29.0 & 32.0 & 30.0 & 44.0 & 54.0 & 46.0 & 73.1 & 66.6 & 65.8 & 63.6 & 33.7 & 72.0 & 45.3 \\
\textbf{-8x7b} & \textbf{One-shot} & 35.2 & 35.0 & 32.2 & 31.0 & 35.0 & 29.1 & 29.0 & 32.0 & 30.0 & 36.5 & 47.1 & 46.0 & 50.1 & 53.1 & 51.2 & 51.1 & 23.6 & 61.5 & 39.4 \\
\textbf{-Instruct} & \textbf{DS-Agent} & 37.0 & 35.0 & 35.0 & 31.0 & 35.0 & 32.0 & 29.0 & 32.0 & 30.0 & {\ul 20.1} & 16.4 & 38.5 & {\ul 25.3} & 54.5 & 53.7 & 53.9 & 32.2 & 47.6 & 35.5 \\ \midrule
\multirow{3}{*}{\textbf{GPT-3.5}} & \textbf{Zero-shot} & 21.7 & 35.0 & 30.1 & 28.6 & 27.1 & 28.3 & 27.1 & 29.1 & 28.1 & 33.1 & 48.4 & 21.4 & 29.0 & 35.3 & 28.8 & 35.7 & 25.2 & 42.3 & 30.8 \\
 & \textbf{One-shot} & 27.6 & 25.8 & 27.6 & 25.6 & 34.6 & 23.0 & 20.8 & 29.1 & 27.0 & 35.7 & 48.4 & \textbf{21.1} & 27.1 & 50.5 & 58.4 & 57.5 & 33.9 & 56.4 & 35.0 \\
 & \textbf{DS-Agent} & \textbf{6.0} & {\ul 22.6} & 15.0 & {\ul 20.6} & {\ul 15.1} & \textbf{13.1} & {\ul 17.3} & {\ul 13.4} & {\ul 14.4} & \textbf{20.0} & {\ul 13.0} & 23.0 & 29.0 & {\ul 19.3} & \textbf{7.6} & \textbf{2.0} & 37.0 & {\ul 19.5} & {\ul 17.1} \\ \midrule
\multirow{3}{*}{\textbf{GPT-4}} & \textbf{Zero-shot} & 36.7 & 31.8 & 35.0 & 29.0 & 29.4 & 32.0 & 29.0 & 32.0 & 30.0 & 37.3 & 45.7 & 33.6 & \textbf{1.0} & \textbf{15.3} & 23.2 & 17.9 & 28.3 & 20.1 & 28.2 \\
 & \textbf{One-shot} & 35.1 & 24.4 & \textbf{13.8} & 26.6 & 29.6 & 28.8 & 23.1 & 30.1 & 26.6 & 26.7 & 41.6 & 36.7 & 29.7 & 21.9 & 35.3 & 28.9 & {\ul 21.4} & 23.2 & 28.0 \\
 & \textbf{DS-Agent} & {\ul 18.6} & \textbf{1.0} & {\ul 14.6} & \textbf{5.2} & \textbf{6.2} & {\ul 18.8} & \textbf{15.7} & \textbf{6.3} & \textbf{8.1} & \textbf{20.0} & \textbf{11.4} & {\ul 21.2} & \textbf{1.0} & 32.6 & {\ul 14.5} & {\ul 8.2} & \textbf{13.0} & \textbf{12.4} & \textbf{12.7} \\ \bottomrule
\end{tabular}
}
\label{tab:main-result-stage-2}
\vspace{-15pt}
\end{table*}

\subsubsection{Ablation Study}
To validate the effectiveness of the CBR paradigm in the development stage, we conduct two ablation studies on DS-Agent, and the results are presented in Table \ref{tab:ablation-stage-1}.

Firstly, we investigate \textbf{(1) w/o ReviseRank}, which directly utilizes the top-ranked retrieved case without adjusting the retrieved case based on execution feedback, which can be also regarded as a RAG based LLM agent. As expected, this ablation leads to performance deterioration, indicating the importance of adjusting the retrieved case based on the execution feedback in the retrieval process.

Next, we evaluate \textbf{(2) w/o CBR} to verify the overall effectiveness of the CBR paradigm, which prompts LLMs to generate experiment plans without incorporating human insights. This variant yields the worst performance among three agents, since LLMs are not trained to align with the data science scenario, thus incapable of autonomously formulating reasonable experiment plans. Integrating the CBR paradigm successfully addresses this limitation, empowering LLMs to adeptly incorporate the expert knowledge from Kaggle to solve data science tasks.

\begin{figure}[t]
    \begin{center}
    \centerline{\includegraphics[width=0.8\linewidth]{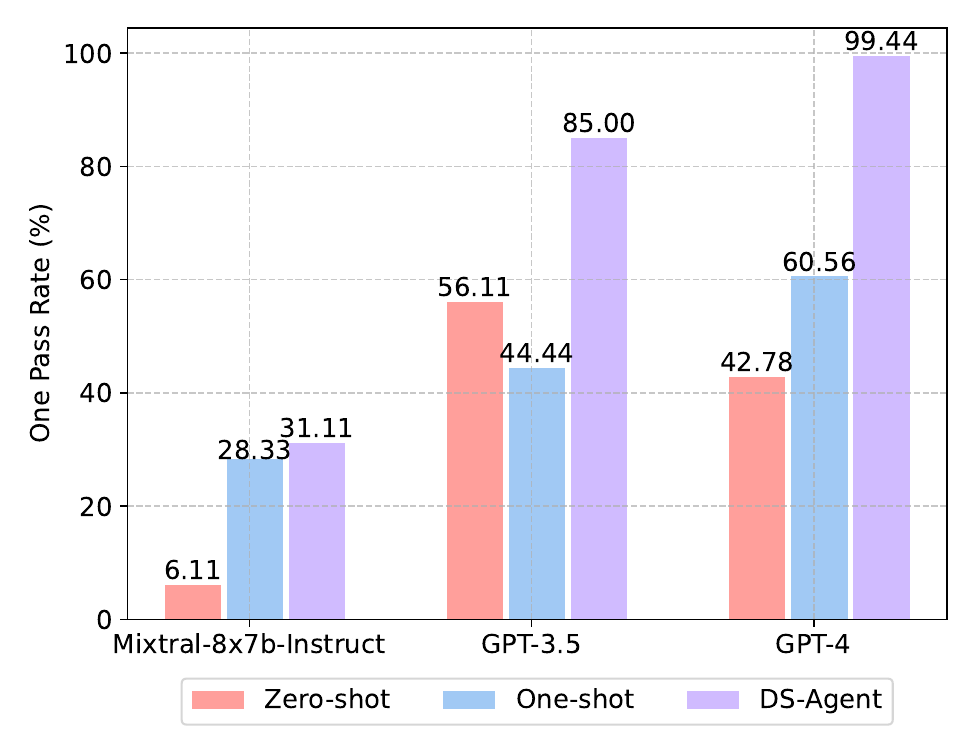}}
    \caption{One pass rate of nine different agents over 18 deployment tasks. The reported results are averaged across 10 random runs.}
    \label{fig:one-pass-rate-stage-2}
    \end{center}
    \vspace{-20pt}
\end{figure}

\subsection{Results for Deployment Stage}
\subsubsection{Main Results}
\noindent\textbf{Baselines.} In the deployment stage, we compare DS-Agent with two baselines: \textbf{(1) Zero-shot} directly prompts LLMs for code generation. \textbf{(2) One-shot} incorporates a random example case from the agent case bank into the context of LLMs. This can also be considered as an ablation of the retrieval process. All agents are implemented using GPT-3.5, GPT-4, and an open-source LLM \texttt{Mixtral-8x7b-Instruct} \cite{mixtral}.

\begin{figure*}[t]
  \centering
  \subfigure[]{\includegraphics[width=0.77\linewidth]{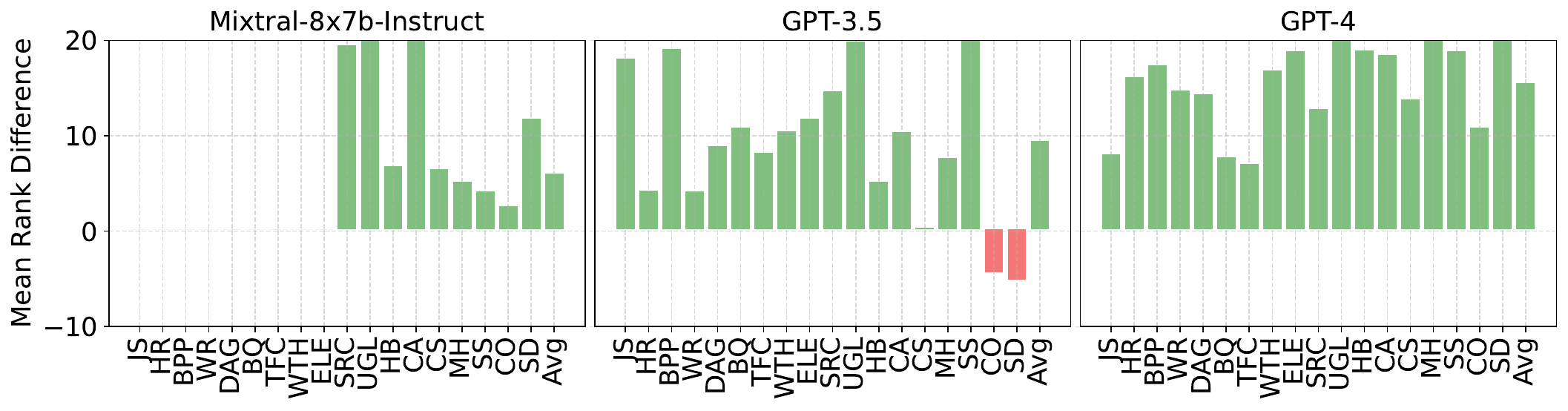}}
  \vspace{-10pt}
  \subfigure[]{\includegraphics[width=0.215\linewidth]{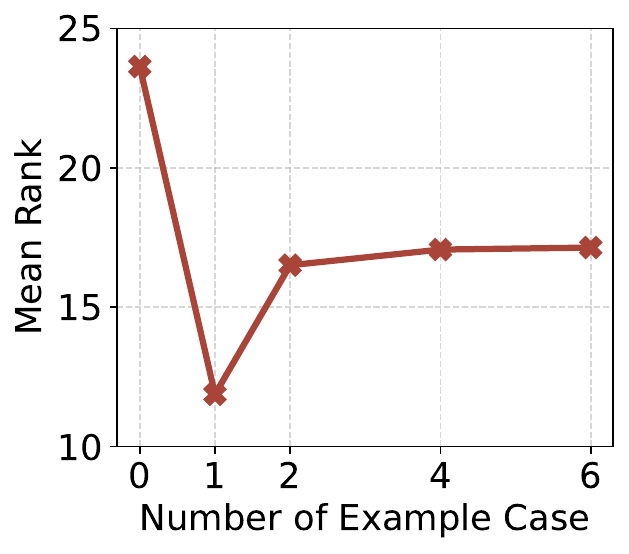}}
  \caption{Further analyses on DS-Agent in the deployment stage. (a) Performance difference of DS-Agent learning from past successful experiences or textual human insights. (b) Hyper-parameter study on varying number of example case in DS-Agent with GPT-3.5.}
  \label{fig:merge-ablation-parameter}
  \vspace{-10pt}
\end{figure*}

\noindent\textbf{Comparison on one pass rate.} First, we investigate the one pass rate of nine different agents over 18 deployment tasks. As depicted in Figure \ref{fig:one-pass-rate-stage-2}, DS-Agent demonstrates remarkable superiority over alternative baselines across various LLMs. Particularly noteworthy is the outstanding performance of DS-Agent with GPT-4, achieving an unprecedented one pass rate of nearly 100\%. Moreover, DS-Agent with GPT-3.5, attains the second-highest one pass rate at 85\%. Furthermore, DS-Agent with Mixtral-8x7b-Instruct results in a notable 25\% improvement in one pass rate compared to the zero-shot strategy. These results highlight the efficacy of the CBR paradigm in augmenting the bug-free programming capabilities of LLMs for data science tasks. While the one-shot strategy generally brings improvement compared to the zero-shot approach, there is an exception in the case of GPT-3.5, possibly attributed to its comparatively inferior reasoning capabilities. Additionally, DS-Agent consistently outperforms one-shot strategy, underscoring the importance of the retrieval process.

\noindent\textbf{Comparison on task-specific evaluation metric.}
Then, our focus shifts to the task-specific performance of 18 deployment tasks, as outlined in Table \ref{tab:main-result-stage-2}. Notably, DS-Agent with GPT-4 attains the highest mean rank among the nine agents, while DS-Agent with GPT-3.5 secures the second-highest mean rank, even surpassing baselines with GPT-4. Unfortunately, DS-Agent with the open-source LLM still exhibits weaker performance than agents with GPT-3.5 or GPT-4, attributable to its inferior foundational LLM capabilities. Nevertheless, it still outperforms or competes favorably with other baselines with the open-sourced LLM in 13 out of 18 deployment tasks. These empirical observations substantiate the efficacy of the proposed CBR paradigm.

\noindent\textbf{Comparison on resource cost.}
One crucial design of DS-Agent lies in its two distinct stages. The development stage focuses on exploring effective model deigns, incurring relatively high resource cost, while the deployment stage is tailored for swiftly and efficiently solving data science tasks with minimal resources. As shown in Table \ref{tab:cost-stage-1-2}, in the deployment stage, DS-Agent incurs costs of \$0.0045 and \$0.1350 for a single run with GPT-3.5 and GPT-4, respectively. This represents a substantial cost reduction of over 90\% when compared to the development stage, rendering DS-Agent highly appealing for real-world deployment scenarios.

\begin{table}[t]
\caption{Monetary cost comparison among development and deployment stage on a single run.}
\centering
\resizebox{0.9\linewidth}{!}{
\begin{tabular}{@{}lccc@{}}
\toprule
\multirow{2}{*}{\textbf{DS-Agent}} & \textbf{Development} & \textbf{Deployment} & \textbf{Cost Deduction} \\
 & \textbf{Stage} & \textbf{Stage} & \textbf{Percentage} \\ \midrule
\textbf{GPT-3.5} & \$0.06 & \$0.0045 & 92.5\% \\
\textbf{GPT-4} & \$1.60 & \$0.1350 & 91.5\% \\ \bottomrule
\end{tabular}
}
\label{tab:cost-stage-1-2}
\end{table}

\subsubsection{Further Analyses}
\label{sec:stage2-analyses}
\noindent\textbf{Ablation study.}
In the development stage, DS-Agent adapts past successful agent experiences to solve the unseen data science tasks. One natural idea is to directly integrate the collected textual human insights from the development stage into the context of LLMs to enhance its data science capabilities. To this end, we investigate an ablation variant of DS-Agent, which learns from relevant human insight for code generation in the deployment stage. As shown in Figure \ref{fig:merge-ablation-parameter}~(a), DS-Agent that learns from past successful experiences significantly outperforms its counterpart that learns from textual human insights across nearly all tasks. This demonstrates that learning from the homogeneous case (i.e., an example task and one of its solutions) leads to better performance than from the heterogeneous case (i.e., textual solution insights). This finding emphasizes the crucial role of both development and deployment stages in DS-Agent.

\noindent\textbf{Hyper-parameter analysis on case number in the context.} Next, we delve into a crucial hyperparameter within DS-Agent: the number of retrieved example cases in the context of LLMs, as illustrated in Figure \ref{fig:merge-ablation-parameter}~(b). Notably, DS-Agent, when devoid of example cases, regresses to the zero-shot strategy, resulting in the poorest performance among all the settings. This underscores the ability of LLMs to glean valuable insights from contextual cases for solving ML tasks. Intriguingly, as the number of example cases in the context increases, the performance of DS-Agent experiences a rapid decline—an unexpected outcome in typical few-shot learning scenarios. It is essential to highlight that the reuse process in DS-Agent is centered around adapting a single example case to address the current ML task. Therefore, the presence of more than one example case in the context introduces interference information to LLMs, hindering its ability to generate appropriate code for the current task.

\section{Related Work}

\noindent\textbf{LLM Agent.}
LLMs have demonstrated remarkable foundational capabilities, such as language understanding, complex reasoning, tool usage, and code generation, which gives rise to the development of autonomous language agents designed for various tasks \cite{react,metagpt,autogen,deps,expel,chemical-research,mathematical-discoveriy,Deng2023K2AF,Lin2023GeoGalacticaAS}. Within the field of data science, \citet{chatgpt-ds} discusses the potential of LLMs as conversational agents in data science workflows. Moreover, recent studies have investigated the use of LLM agents in diverse areas such as feature engineering \cite{caafe}, hyper-parameter tuning \cite{automl-gpt, mlcopilot}, using ML libraries \cite{ml-bench}, aiding AI research \cite{mlagentbench}, data operation \cite{ds-1000}, etc. In contrast to them, we focus on developing automatic language agents to build and train ML models, contributing to the field of automated data science. Concurrent to our work, \citet{DI} propose Data Interpreter, which focuses on optimizing the workflow of LLM agents for data science scenarios to fully unlock the intrinsic knowledge of LLMs. The core techniques of Data Interpreter and DS-Agent are complementary. A potential future work would be to enhance DS-Agent with Data Interpreter or integrate CBR into Data Interpreter for further improvement.

\noindent\textbf{Case-Based Reasoning.}
Case-Based Reasoning (CBR) \cite{cbr-review-1, cbr-review-2, cbr-review-3}, a classical AI paradigm proposed several decades ago, aims to address new problems by adapting insights from analyzing and reasoning with retrieved relevant cases. Integrating CBR into LLMs \cite{cbr-1,cbr-2,cbr-3} exhibits procedural similarities to the well-known retrieval-augmented generation (RAG) framework \cite{rag-kinlp, rag-1, rag-2, rag-survey}, particularly in the steps of retrieval and reuse. However, a distinctive feature of CBR lies in its feedback mechanism, which enables iteratively adjusting the retrieved cases and revising the solutions accordingly. Furthermore, CBR enhances future problem-solving by retaining and reusing successful cases.
\section{Conclusion}
In this work, we propose DS-Agent, a novel framework that harnesses LLM agent and case-based reasoning to solve data science tasks. In the development stage, DS-Agent structures an automatic iteration pipeline on top of the CBR framework, which aims to retrieve and reuse relevant human insights from Kaggle to develop the experiment plan, and then iteratively adjust the retrieved cases and revise the plan based on the execution feedback. As for the deployment stage, DS-Agent leverages a simplified CBR framework to achieve a low-resource scenario by retrieving and reusing the successful solution cases collected from the development stage. Extensive experiments are conducted to demonstrate the effectiveness of DS-Agent for data science tasks.

\section*{Acknowledgements}
We truly thank the reviewers for their great effort in our submission. This work was supported by the National Key R\&D Program of China under Grant (No. 2023YFF0905400), National Natural Science Foundation of China through grants (No. U2341229, No. 61976102, No. U19A2065), the Key R\&D Project of Jilin Province, China, (No. 20240304200SF), and the International Cooperation Project of Jilin Province, China, (No. 20220402009GH).

\section*{Impact Statement}
Here we emphasize some potential ethics concerns of DS-Agent: \textbf{(1) Unemployment and skill obsolescence.} The major concern of our research lies in potential unemployment and skill obsolescence. However, as discussed in \cite{automl-survey}, the intent of automated data science is not to replace data scientists but rather to assist them, allowing them to concentrate on more complex aspects of data science work. Wherein they only need to focus on higher-level data science problems, such as task formulation, data visualization, cleaning and curation, prediction engineering, and result summary and recommendation. Moreover, by enabling interaction through natural language, automated data science lowers the barrier to entry, facilitating a more accessible pathway for users to glean insights from data, thereby democratizing the field of data science. \textbf{(2) Malicious code generation.} An often underappreciated yet critical concern with the proliferation of automated data science tools like DS-Agent is the potential for generating code that could be detrimental to computational devices or data integrity. As DS-Agent navigates the vast terrain of possible solutions to a given data problem, it may inadvertently produce code that is inefficient, vulnerable to exploitation, or even directly harmful. While such issues were not observed in our experiments, it is prudent for users to review any code produced by DS-Agent before execution. To enhance security, we recommend running DS-Agent within a Docker container, which provides a layer of isolation for the host's file system. \textbf{(3) Data Privacy and Security.} To protect data privacy and security, DS-Agent is designed to operate locally, negating the need to upload sensitive data. However, when integrating API-based Large Language Models (LLMs) like GPT-3.5 or GPT-4, there is an inherent privacy risk since these interactions typically involve transmitting data to external servers. We advise users to carefully inspect any data sent in API prompts to prevent unintentional data disclosures.


\bibliography{references}
\bibliographystyle{icml2024}

\newpage
\appendix
\onecolumn
\section*{Appendix}
The appendix of this paper is organized as below. First, we provide the pseudo-code of DS-Agent in \autoref{app:alg}. We then provide experimental details, including task selection (Appendix \ref{app:task}), details for human insight collection (Appendix \ref{app:human-insight}), and model configuration and hyper-parameter settings (Appendix \ref{app:setting}). Subsequently, we provide further discussions on AutoML techniques (Appendix \ref{app:automl}). Moreover, we provide two case studies of DS-Agent in Appendix \ref{app:case-study} and detailed error mode analyses in Appendix \ref{app:ema}. Finally, the detailed prompt design in DS-Agent is presented in \autoref{app:prompt}.

\section{Pseudo-code of DS-Agent}
\label{app:alg}
We present the pseudo-code for DS-Agent in two stages: the development stage (Algorithm \ref{alg:development}) and the deployment stage (Algorithm \ref{alg:deployment}).

\begin{algorithm}[h]
   \caption{Development Stage of DS-Agent}
   \label{alg:development}
\begin{algorithmic}[1]
\STATE {\bfseries Initialization:} Development task set $\mathcal{T}_\text{develop}$, Human insight case bank $\mathcal{C}$, Agent experience case bank $\mathcal{B}=\varnothing$, Embedding model $\textbf{E}(\cdot)$, number of the retrieved cases $k$, ReviseRank agent \texttt{ReviseRanker}, Planning agent \texttt{Planner}, Programming agent \texttt{Programmer}, Debugging agent \texttt{Debugger}, Logging agent \texttt{Logger}.
\FOR {$\tau$ {\bfseries in} $\mathcal{T}_\text{develop}$}
\STATE Initialize experiment log $l^0=\{\}$
\STATE Retrieve cases $c_1, c_2, ..., c_k$ with top-$k$ cosine similarities from the human insight case bank $\mathcal{C}$
\FOR {$t$ {\bfseries in} $1,2,...,T$}
\STATE Revise the ranking order of $c_1, c_2, ..., c_k$ using $\texttt{ReviseRanker}(c_1, c_2, ..., c_k, \tau, l^{t-1})$ 
\STATE Select the top-ranked case as $c^t$
\STATE Reuse $c^t$ to develop the experiment plan $y^t$ with $\texttt{Planner}(c^t, \tau, l^{t-1})$
\STATE Generate Python code $s^t$ based on the experiment plan $y^t$ using $\texttt{Programmer}(\tau, s^{t-1}, y^t)$
\STATE Execute the code $s^t$ and observe the execution result $o^t$
\WHILE{there are errors reported in $o^t$ \AND number of debugging $n_{\text{debug}} < N$}
\STATE Debug and generate the corrected code $s^t$ using $\texttt{Debugger}(\tau, s^{t-1}, y^t, s^t, o^t)$
\STATE Execute the corrected code $s^t$ and observe the execution result $o^t$
\ENDWHILE
\STATE Write the experiment log $l^t$ with $\texttt{Logger}(\tau, l^{t-1}, y^t, s^{t-1}, s^t, o^t)$
\IF {the performance on test set is improved}
\STATE Store the task description and code $\mathcal{B} \leftarrow (\tau, s^t)$.
\ENDIF
\ENDFOR
\ENDFOR
\end{algorithmic}
\end{algorithm}

\begin{algorithm}[h]
   \caption{Deployment Stage of DS-Agent}
   \label{alg:deployment}
\begin{algorithmic}[1]
\STATE {\bfseries Initialization:} Deployment task set $\mathcal{T}_\text{deploy}$, Agent case bank $\mathcal{B}$, Embedding model $E(\cdot)$, Adaptation agent $\texttt{Adapter}$.
\FOR {$\tau$ {\bfseries in} $\mathcal{T}_\text{deploy}$}
\STATE Retrieve past example case $(\tau_0, s_0))$ from $\mathcal{B}$ with top-ranked similarity
\STATE Generate code $s$ using $\texttt{Adapter}(\tau_0, s_0, \tau)$
\ENDFOR
\end{algorithmic}
\end{algorithm}

\section{Experimental Details}
\label{app:exp-details}
\subsection{Task Selection}
\label{app:task}
We select 30 representative data science tasks covering three data modalities and two fundamental ML task types. The detailed descriptions of these tasks are presented in \autoref{tab:task-description}. We introduce various evaluation metrics for these tasks, including accuracy, area under the receiver operating characteristic curve (AUROC), negative log likelihood (NLL), mean column-wise root mean squared error (MCRMSE), mean squared error (MSE), root mean log squared error (RMLSE), mean absolute error (MAE), root mean squared error (RMSE) and median squared error (MedAE). Notably, the majority of these datasets were released after September 2021, ensuring they were not part of the pretraining corpus for LLMs. For each task, we write a natural language task description, as well as a Python script that establishes a baseline of random guess to serve as an initial reference point for the agents. We showcase an example task of airline-reviews (AR) as follows.
\begin{tcolorbox}[title={\textbf{\small Task Description}}, 
boxrule=2pt, arc=0mm]
{\scriptsize
\begin{lstlisting}[breaklines,basicstyle=\ttfamily,frame=single,]
You are solving this machine learning tasks of regression: 
The dataset presented here (the Airline reviews) comprises customer feedback for British Airways. Here, we provide the textual reviews. Your task is to predict the corresponding rating in the range of {1, ..., 10} given the reviews in the test set. The evaluation metric is root mean squared error (RMSE).
We provide an overall pipeline in train.py. Now fill in the provided train.py script to train a language model to get a good performance.  
\end{lstlisting}
}
\end{tcolorbox}

\begin{tcolorbox}[title={\textbf{\small The provided Python script (train.py)}}, 
boxrule=2pt, arc=0mm]
{\scriptsize
\begin{lstlisting}[breaklines,basicstyle=\ttfamily,frame=single,language=Python,]
import pandas as pd
from sklearn.metrics import mean_squared_error
import numpy as np
import random
import torch
from sklearn.model_selection import train_test_split
from submission import submit_predictions_for_test_set

SEED = 42
random.seed(SEED)
torch.manual_seed(SEED)
np.random.seed(SEED)
device = torch.device("cuda" if torch.cuda.is_available() else "cpu")

def compute_metrics_for_regression(y_test, y_test_pred):
    rmse = mean_squared_error(y_test, y_test_pred, squared=False) 
    return rmse

def train_model(X_train, y_train, X_valid, y_valid):
    # TODO. define and train the model
    # should return the trained model
    model = None
    return model

def predict(model, X):
    # TODO. predict the model
    # should return an array of predictions
    y_pred = np.random.randint(1, 11, len(X))
    return y_pred

if __name__ == '__main__':
    data_df = pd.read_csv('train.csv')
    data_df = data_df.dropna(subset=['OverallRating'])
    
    # Process data and store into numpy arrays.
    X = list(data_df.ReviewBody.to_numpy())
    y = data_df.OverallRating.to_numpy()

    # Create a train-valid split of the data.
    X_train, X_valid, y_train, y_valid = train_test_split(X, y, test_size=0.10, random_state=SEED)

    # define and train the model
    # should fill out the train_model function
    model = train_model(X_train, y_train, X_valid, y_valid)

    # evaluate the model on the valid set using compute_metrics_for_regression and print the results
    # should fill out the predict function
    y_valid_pred = predict(model, X_valid)
    rmse = compute_metrics_for_regression(y_valid, y_valid_pred)
    print("final RMSE on validation set: ", rmse)

    # submit predictions for the test set
    submission_df = pd.read_csv('test.csv')
    submission_df = submission_df.dropna(subset=['OverallRating'])
    X_submission = list(submission_df.ReviewBody.to_numpy())
    y_submission = predict(model, X_submission)
    submit_predictions_for_test_set(y_submission)
  
\end{lstlisting}
}
\end{tcolorbox}

\begin{table}[t]
\caption{Detailed descriptions of selected data science tasks in the experiment.}
\resizebox{\linewidth}{!}{
\begin{tabular}{@{}clclcccccc@{}}
\toprule
\textbf{Stage} & \multicolumn{1}{c}{\textbf{Dataset Name}} & \textbf{Abbr.} & \multicolumn{1}{c}{\textbf{Resource}} & \textbf{Modality} & \textbf{Task} & \textbf{Evaluation Metric} & \textbf{Train} & \textbf{Valid} & \textbf{Test} \\ \midrule
\multirow{12}{*}{\textbf{Development}} & feedback & FB & Kaggle Competition & Text & Regression & MCRMSE & 3449 & 383 & 79 \\
 & airline-reviews & AR & Kaggle Dataset & Text & Regression & RMSE & 2997 & 333 & 371 \\
 & textual-entailment & TE & Kaggle Dataset & Text & Classification & Accuracy & 4417 & 490 & 4908 \\
 & chatgpt-prompt & CP & Kaggle Dataset & Text & Classification & Accuracy & 468 & 116 & 585 \\
 & ett-m2 & ETT & Research Dataset & Time Series & Forecasting & MSE & 34465 & 11521 & 11521 \\
 & ili & ILI & Research Dataset & Time Series & Forecasting & MSE & 617 & 74 & 170 \\
 & handwriting & HW & Research Dataset & Time Series & Classification & Accuracy & 150 & 0 & 850 \\
 & ethanol-concentration & EC & Research Dataset & Time Series & Classification & Accuracy & 261 & 0 & 263 \\
 & media-campaign-cost & MCS & Kaggle Competition & Tabular & Regression & RMLSE & 291872 & 32430 & 324303 \\
 & wild-blueberry-yield & WBY & Kaggle Competition & Tabular & Regression & MAE & 12384 & 1376 & 13761 \\
 & spaceship-titanic & ST & Kaggle Competition & Tabular & Classification & Accuracy & 6259 & 695 & 1739 \\
 & enzyme-substrate & ES & Kaggle Competition & Tabular & Classification & AUROC & 12019 & 1335 & 13355 \\ \midrule
\multirow{18}{*}{\textbf{Deployment}} & jigsaw & JS & Kaggle Dataset & Text & Regression & RMSE & 8639 & 959 & 720 \\
 & bitcoin-price-prediction & BPP & Kaggle Dataset & Text & Regression & RMSE & 1757 & 195 & 217 \\
 & hotel-reviews & HR & Kaggle Dataset & Text & Regression & RMSE & 9220 & 1024 & 1025 \\
 & webmd-reviews & WR & Kaggle Dataset & Text & Classification & Accuracy & 11612 & 2903 & 871 \\
 & detect-ai-generation & DAG & Kaggle Dataset & Text & Classification & Accuracy & 8751 & 2187 & 1093 \\
 & boolq & BQ & Kaggle Dataset & Text & Classification & Accuracy & 1308 & 327 & 1635 \\
 & traffic & TFC & Research Dataset & Time Series & Forecasting & MSE & 12185 & 1757 & 3509 \\
 & weather & WTH & Research Dataset & Time Series & Forecasting & MSE & 36792 & 5271 & 10540 \\
 & electricity & ELE & Research Dataset & Time Series & Forecasting & MSE & 18317 & 2633 & 5261 \\
 & self-regulation-scp1 & SRC & Research Dataset & Time Series & Classification & Accuracy & 268 & 0 & 293 \\
 & uwave-gesture-library & UGL & Research Dataset & Time Series & Classification & Accuracy & 120 & 0 & 320 \\
 & heartbeat & HB & Research Dataset & Time Series & Classification & Accuracy & 204 & 0 & 250 \\
 & crab-age & CA & Kaggle Competition & Tabular & Regression & MAE & 59981 & 6664 & 66646 \\
 & concrete-strength & CS & Kaggle Competition & Tabular & Regression & RMSE & 4380 & 486 & 4867 \\
 & mohs-hardness & MH & Kaggle Competition & Tabular & Regression & MedAE & 8430 & 936 & 9367 \\
 & cirrhosis-outcomes & CO & Kaggle Competition & Tabular & Classification & NLL & 6403 & 711 & 7115 \\
 & smoker-status & SS & Kaggle Competition & Tabular & Classification & AUROC & 128997 & 14333 & 143331 \\
 & software-defects & SD & Kaggle Competition & Tabular & Classification & AUROC & 82428 & 9158 & 91587 \\ \bottomrule
\end{tabular}
}
\label{tab:task-description}
\end{table}

\subsection{Details for Human Insight Case Collection}
\label{app:human-insight}
A primary requirement of DS-Agent in the development stage lies in collecting human insights from Kaggle. Specifically, we select a total of 12 recently completed Kaggle competitions, with four competitions for each data modality, i.e., text, time series, and tabular data. Then, we crawl the technical report shared by the top-10 winner teams in the private leaderboard, and the Jupyter notebooks with top-10 scores in the public leaderboard. For the technical reports, we only perform basic text cleaning to reserve most of the insights. For the codes, we prompt GPT-3.5 to extract the textual solution from them. We present the prompts for processing codes, and two examples of the collected human insight case as below.

\begin{tcolorbox}[title={\textbf{\small Prompt for solution extraction}}, 
boxrule=2pt, arc=0mm]
{\scriptsize
\begin{lstlisting}[breaklines,basicstyle=\ttfamily,frame=single,]
Assume that you were a proficient data scientist. The following Python code is a high-performing solution for a kaggle competition.
Please answer the following questions one by one and **as detailedly as possible**. 
Make sure that another data scientist can exactly reproduce this copy of code based on your answer.
Focus on the training process.
(1) Please give a summary of the overall design. 
(2) What is the overall model architecture? Please use a long article to answer this question as accurately and in detail as possible.
(3) How are the important hyper-parameters setting in this code?
(4) What is the optimization objective?
(5) What advanced machine learning technique does this copy of code use?
(6) What other important tricks do you think play an important role for high performance?
Note that make sure the answers are directly included from the python code, rather than based on your assumption.
```python
{Here is the Python code.}
```
\end{lstlisting}
}
\end{tcolorbox}

\begin{tcolorbox}[title={\textbf{\small An example human insight case that is derived from the public technical report}}, 
boxrule=2pt, arc=0mm]
{\scriptsize
\begin{lstlisting}[breaklines,basicstyle=\ttfamily,frame=single,]
My solution is rather simple, because I used almost no complex tricks - instead I just built a reliable pipeline and found good hyperparameters. I had some brilliant ideas (at least I think so :)) - to use augmentation and generate synthetic data using some big model based on other texts from commonlit.org. But at that time, when I was still actively contributing, there was uncertainty with the license for the texts, and I did not complete the use of augmentations - I was too busy with my work (trying to get RLHF to work), so I left the competition - my last commit was a month ago. But apparently the decision was good, at least I didn't overfit :)

So let's break my solution down into parts.

**1. Data:**
I used a fairly standard template: prompt + question + text. At the end of my participation, I tried to dig into the data and do a good pre-processing - after all, there were quite a lot of very similar essays with exact or different scores. So, I tried to find these samples by similarity (like Levenstein) and then merge them. In addition, I decided to make augumentations based on this insight - if there are many similar essays (differing only in typos, for example) - I could use something like reverse autocorrect - randomly replace some words with its close analogues. With this technique I got 0.453 in private ONLY on fold3 (which is better than my chosen blend and probably could lead to a second place) - but I was too tired at this point so I didn't look further into augmentations. But I think augmentations could probably lead me to victory.

**2. Models**
Deberta is the king, so there's not much to say here. I tried using decoder models like Llama, but Deberta was still better. There were some techniques that gave me a boost - using EMA (honestly, without EMA it was very unstable, so it's probably just necessary) and using differential learning rates.I tried several pooling options, but the best option for me was to use concatenation of CLS token and student's text meanpooling. I also used token_type_ids to separate the prompt, question and essay.

**3 Inference & train**
I used following scheme - I tried to find a good hyperparameters on some fold (for example, fold0), and then train with exact hyperparameters on other folds. I then sumbittend the entire blend and 4 individual models (5 submission total - one day) and repeated the procedure the next day. I realized that I could use maxlen 1500 for inference (didn't research this number much, tried something like 1024 and 2048, but 1500 was better in terms of efficiency), so in my final mix I took the 10 best checkpoints across folds (some folds got 2 checkpoints, some folds got 3). First I averaged by folds, then averaged the rest. That's all. 

Briefly what worked (ranked from most important to least important, IMO):
1. Using Deberta
2. EMA
3. Augumentation
4. Defferentiated learning rates
5. Custom pooling
6. token_type_ids
7. Data cleaninig

What did not work (random order):
1. Decoder models
2. AWP
3. FGM
4. WD
5. Constant LR
6. Handcrafted features
7. GBT for stacking

In the end, it was a good competition for me. Last year I competed in another NLP competition and got a silver medal, but I grinded all day at that competition (I wasn't working that time, so I had a lot of free time). This time I also expected silver, which I consider a solid result, but I got 3rd place. In any case, this competition was a cakewalk for me, since I spend very little effort on it (compared to the previous competition, at least). I'm hoping this means I'll grow a lot this year - and I think that's the main goal of participating in Kaggle.

Good luck to all of you.


\end{lstlisting}
}
\end{tcolorbox}

\begin{tcolorbox}[title={\textbf{\small An example human insight case that is derived from the public codes}}, 
boxrule=2pt, arc=0mm]
{\scriptsize
\begin{lstlisting}[breaklines,basicstyle=\ttfamily,frame=single,]
(1) The overall design of the code is to train a DebertaV3 model for predicting the content and wording scores of student summaries. The code includes data preprocessing, model training, validation, and prediction steps.

(2) The overall model architecture is based on the DebertaV3 model, which is a transformer-based model. The code uses the `AutoModelForSequenceClassification` class from the `transformers` library to load the pre-trained DebertaV3 model. The model is fine-tuned for sequence classification with a single output label. The input to the model is a concatenation of the prompt question, summary text, and prompt text. The model tokenizes the input using the `AutoTokenizer` class and generates input tensors for the model. The model architecture consists of multiple transformer layers with self-attention mechanisms, followed by a linear layer for classification.

(3) The important hyperparameters in this code are set in the `CFG` class. The hyperparameters include the model name, learning rate, weight decay, hidden dropout probability, attention dropout probability, number of training epochs, number of cross-validation splits, batch size, random seed, save steps, and maximum sequence length.

(4) The optimization objective is to minimize the root mean squared error (RMSE) between the predicted scores and the ground truth scores. The code uses the mean squared error (MSE) as the loss function and calculates the RMSE as the evaluation metric.

(5) The advanced machine learning technique used in this code is transfer learning. The code loads a pre-trained DebertaV3 model and fine-tunes it on the student summary dataset. Transfer learning allows the model to leverage knowledge learned from a large pre-training dataset to improve performance on a specific task.

(6) Some important tricks that play a role in high performance include:
- Data preprocessing: The code preprocesses the input data by tokenizing the text, removing stop words, fixing misspellings, and extracting features such as text length, word overlap, n-gram co-occurrence, quotes overlap, and grammar check.
- Model architecture: The code uses the DebertaV3 model, which is a state-of-the-art transformer-based model known for its strong performance on various natural language processing tasks.
- Training strategy: The code uses k-fold cross-validation to train and validate the model on multiple subsets of the data. This helps to reduce overfitting and obtain a more robust evaluation of the model's performance.
- Evaluation metric: The code uses the root mean squared error (RMSE) as the evaluation metric, which is a common metric for regression tasks. This metric penalizes large errors more than mean absolute error (MAE) and provides a more comprehensive measure of the model's performance.
- Feature engineering: The code incorporates additional features such as word difficulty, readability scores, and cosine similarity between the summary and prompt text. These features capture different aspects of the text and can provide additional information for the model to make predictions.
- Ensemble learning: The code combines the predictions from multiple folds of the cross-validation to obtain a more robust prediction. This helps to reduce the variance and improve the overall performance of the model.
\end{lstlisting}
}
\end{tcolorbox}

\subsection{Model Configuration and Hyper-parameter Setting}
\label{app:setting}
For GPT-3.5 and GPT-4, we use the \texttt{gpt-3.5-turbo-16k} and \texttt{gpt-4-0613} models via the OpenAI API. For the open-source LLM, we utilize \texttt{Mixtral-8x7B-Instruct-v0.1} and use the vLLM framework \cite{vllm} for speedup. In the development stage, we use the decoding strategy with temperature $T=0.5$, while we adjust it to $T=0.7$ in the deployment stage to enhance the diversity of generation. We utilize \texttt{llm-embedder} \cite{llm-embedder} as the pretrained embedding language model.

For DS-Agent in the development stage, we set the iteration times $T=5$, number of the retrieved cases $k=5$ and the number of the debugging $n_{\text{debug}}=5$. For the baseline reproduction, we strictly follow the original reported hyper-parameter of ResearchAgent \cite{mlagentbench} to achieve fair comparison.

\section{Further Discussions}
\subsection{Comparison with AutoML Techniques}
\label{app:automl}
Automated machine learning (AutoML) \cite{automl-book, automl-survey} shares a similar objective to DS-Agent, which aims at optimizing the machine learning in the data science workflow. In contrast, DS-Agent benefits from the LLMs and CBR in three aspects as follows. 

Firstly, while AutoML systems often demand extensive domain expertise, software development, and frequent updates to manage various data modalities using the latest ML techniques, DS-Agent can efficiently address data science tasks by simply collecting updated public technical reports and codes from Kaggle. 

Secondly, DS-Agent provides high flexibility by dynamically building and training ML models to solve various kinds of data science tasks. In contrast, most existing AutoML systems typically constrain the task types within the tabular data setting \cite{autogluon-tabular, h2o}. For example, enzyme-substrate (ES) is a multi-task classification task with tabular data. However, the advanced AutoML system AutoGluon \cite{autogluon-tabular} does not naturally support this setting, and instead, users need to re-frame the multi-task classification task into multiple single-task classification tasks to make it compatible with AutoGluon.

Thirdly, DS-Agent revolutionizes user interaction by leveraging a conversational interface, allowing users to describe their data science tasks in natural language. This contrasts sharply with traditional AutoML systems, which require users to engage through code, necessitating a comprehensive understanding of machine learning tasks, objective functions, and optimization strategies \cite{chatgpt-ds}. DS-Agent's intuitive approach not only simplifies the user experience but also makes advanced data science accessible to a broader audience, democratizing the field by removing the barrier of technical expertise.

We now give an empirical comparison between DS-Agent and an advanced AutoML system AutoGluon \cite{autogluon-tabular} in four development tasks. Note that we only include tabular tasks since AutoGluon does not flexibly handle other tasks, such as time series classification and text regression. The experimental results are presented in \autoref{tab:automl}. We repeat five runs for DS-Agent and report both the best result and average result, where DS-Agent with GPT-3.5 has an failed run in the spaceship-titanic task and thus we do not report the average performance in that setting. For AutoGluon, since it does not involve randomness in its system, thus we only report its performance for a single run. As shown in \autoref{tab:automl}, DS-Agent with GPT-4 outperforms AutoGluon by a large margin in 2 out of 4 tabular tasks,  while delivering performance on par with AutoGluon for the other two tasks. This demonstrates the superiority of the proposed DS-Agent.

\begin{table}[t]
\caption{Comparison between DS-Agent and AutoGluon in four tabular development tasks.}
\resizebox{\linewidth}{!}{
\begin{tabular}{@{}clcccc@{}}
\toprule \midrule 
 & \multicolumn{1}{c}{} & \textbf{media-campaign-cost} & \textbf{wild-blueberry-yield} & \textbf{spaceship-titanic} & \textbf{enzyme-substrate} \\ \midrule 
\multicolumn{2}{c}{\textbf{Evaluation   Metric}} & \textbf{RMLSE ($\downarrow$)} & \textbf{MAE ($\downarrow$)} & \textbf{Accuracy ($\uparrow$)} & \textbf{AUROC ($\uparrow$)} \\ \midrule 
\multicolumn{2}{c}{\textbf{AutoGluon}} & 0.2707 & 305 & 0.8044 & 0.8683 \\ \midrule 
\multirow{2}{*}{\textbf{Avg of 5}} & \textbf{DS-Agent (GPT-3.5)} & 0.2702 & 291 & / & 0.5534 \\
 & \textbf{DS-Agent (GPT-4)} & 0.2947 & 267 & 0.7977 & 0.8322 \\ \midrule 
\multirow{2}{*}{\textbf{Best of 5}} & \textbf{DS-Agent (GPT-3.5)} & \textbf{0.1968} & \textbf{180} & 0.8027 & 0.7031 \\
 & \textbf{DS-Agent (GPT-4)} & 0.2858 & 182 & \textbf{0.8062} & \textbf{0.9761} \\  \midrule \bottomrule
\end{tabular}
}
\label{tab:automl}
\end{table}

\subsection{Case Study}
\label{app:case-study}
In this subsection, we showcase two representative cases of DS-Agent for development and deployment tasks.

\subsubsection{Case Study in Development Stage}
Our case study begins by detailing how DS-Agent approaches the chatgpt-prompt task during the developmental phase. Initially, DS-Agent build a simple baseline model utilizing a TF-IDF vectorizer and logistic regression, which yields an accuracy of 75.21\% on the validation set. Progressing to the second step, DS-Agent employs a pretrained BERT language model for classification, marginally enhancing accuracy to 76.06\%. However, a subsequent switch to another pretrained language model, RoBERTa, in the third step results in a decline in performance, with accuracy plummeting to 47.86\%. To address this, DS-Agent, in the fourth step, fine-tunes the learning rate and the number of training epochs, which substantially boosts accuracy to 90.59\%. In the final step, DS-Agent turns to the "large" version of RoBERTa from the "base" version and re-calibrates certain hyper-parameters, culminating in a peak accuracy of 98.46\%.

\begin{lstlisting}[breaklines,basicstyle=\ttfamily\scriptsize,frame=single,]

======Step 1======

[Experiment Plan]
For the next experiment, define a simple baseline model for the text classification task. This can be a logistic regression or a simple neural network. Implement the model definition in the `train_model` function in the python code. Train the model using the training data and evaluate its performance using the validation data. Report the accuracy of the model on the validation data.

[Python Script]
```python
import pandas as pd
from sklearn.metrics import accuracy_score
from sklearn.linear_model import LogisticRegression
import numpy as np
import random
from sklearn.feature_extraction.text import CountVectorizer
from sklearn.model_selection import train_test_split
from submission import submit_predictions_for_test_set

SEED = 42
LABEL_NUM = 8

random.seed(SEED)
np.random.seed(SEED)

device = "cpu"

def compute_metrics_for_classification(y_test, y_test_pred):
    acc = accuracy_score(y_test, y_test_pred) 
    return acc

def train_model(X_train, y_train):
    # define a CountVectorizer to convert text into a matrix of token counts
    vectorizer = CountVectorizer()
    X_train_counts = vectorizer.fit_transform(X_train)

    # define the model
    model = LogisticRegression(random_state=SEED).fit(X_train_counts, y_train)
    
    return model, vectorizer

def predict(model, vectorizer, X):
    # convert text into a matrix of token counts
    X_counts = vectorizer.transform(X)
    
    # predict the model
    y_pred = model.predict(X_counts)
    return y_pred

if __name__ == '__main__':
    data_df = pd.read_csv('train.csv')
    
    # Process data and store into numpy arrays.
    X = data_df.Sentence.to_numpy()
    y = data_df.Label.to_numpy()

    # Create a train-valid split of the data.
    X_train, X_valid, y_train, y_valid = train_test_split(X, y, test_size=0.20, random_state=SEED)

    # define and train the model
    model, vectorizer = train_model(X_train, y_train)

    # evaluate the model on the valid set using compute_metrics_for_classification and print the results
    y_valid_pred = predict(model, vectorizer, X_valid)
    acc = compute_metrics_for_classification(y_valid, y_valid_pred)
    print("final Accuracy on validation set: ", acc)

    # submit predictions for the test set
    submission_df = pd.read_csv('test.csv')
    X_submission = submission_df.Sentence.to_numpy()
    y_submission = predict(model, vectorizer, X_submission)
    submit_predictions_for_test_set(y_submission)
```

[Result]
final Accuracy on validation set:  0.7521367521367521
final Accuracy on test set:  0.9145299145299145

[Experiment Log]
Experiment Summary: In the last step, a simple baseline model for text classification was defined using logistic regression. The `train_model` function was modified to include a `CountVectorizer` to convert the text into a matrix of token counts. The model was trained using the training data and evaluated on the validation data.
Experiment Result: The accuracy of the model on the validation data was 0.7521367521367521. 

======Step 2======

[Experiment Plan]
The next step in the experiment will involve implementing a transformer-based model, such as BERT, for the text classification task. The input text will be tokenized and passed through the transformer model. The output of the transformer model will be fed into a linear layer for classification. The learning rate and the number of training epochs will be tuned to optimize the performance of the model. The performance of the model will be evaluated using the accuracy metric on the validation set.

[Python Script]
```python
import pandas as pd
from sklearn.metrics import accuracy_score
from sklearn.model_selection import train_test_split
from transformers import BertTokenizer, BertForSequenceClassification, AdamW
from torch.utils.data import DataLoader, TensorDataset
import torch
import numpy as np
import random
from submission import submit_predictions_for_test_set

SEED = 42
LABEL_NUM = 8
LEARNING_RATE = 2e-5
EPOCHS = 4

random.seed(SEED)
np.random.seed(SEED)
torch.manual_seed(SEED)

device = torch.device("cuda" if torch.cuda.is_available() else "cpu")

def compute_metrics_for_classification(y_test, y_test_pred):
    acc = accuracy_score(y_test, y_test_pred) 
    return acc

def train_model(dataloader, model, optimizer):
    model.train()
    for epoch in range(EPOCHS):
        for batch in dataloader:
            b_input_ids, b_labels = batch[0].to(device), batch[1].to(device)
            model.zero_grad()
            outputs = model(b_input_ids, labels=b_labels)
            loss = outputs[0]
            loss.backward()
            optimizer.step()
    return model

def predict(model, dataloader):
    model.eval()
    predictions = []
    for batch in dataloader:
        b_input_ids = batch[0].to(device)
        with torch.no_grad():
            outputs = model(b_input_ids)
        logits = outputs[0]
        logits = logits.detach().cpu().numpy()
        predictions.extend(np.argmax(logits, axis=1).flatten())
    return predictions

if __name__ == '__main__':
    data_df = pd.read_csv('train.csv')
    
    # Process data and store into numpy arrays.
    X = data_df.Sentence.to_numpy()
    y = data_df.Label.to_numpy()

    # Create a train-valid split of the data.
    X_train, X_valid, y_train, y_valid = train_test_split(X, y, test_size=0.20, random_state=SEED)

    # define the tokenizer and model
    tokenizer = BertTokenizer.from_pretrained('bert-base-uncased', do_lower_case=True)
    model = BertForSequenceClassification.from_pretrained('bert-base-uncased', num_labels=LABEL_NUM)
    model.to(device)

    # tokenize and encode sequences for the training and validation sets
    encoded_data_train = tokenizer.batch_encode_plus(
        X_train.tolist(), 
        add_special_tokens=True, 
        return_attention_mask=False, 
        pad_to_max_length=True, 
        max_length=256, 
        return_tensors='pt'
    )

    encoded_data_valid = tokenizer.batch_encode_plus(
        X_valid.tolist(), 
        add_special_tokens=True, 
        return_attention_mask=False, 
        pad_to_max_length=True, 
        max_length=256, 
        return_tensors='pt'
    )

    # create Tensor datasets
    dataset_train = TensorDataset(encoded_data_train['input_ids'], torch.tensor(y_train))
    dataset_valid = TensorDataset(encoded_data_valid['input_ids'], torch.tensor(y_valid))

    # create dataloaders
    dataloader_train = DataLoader(dataset_train, batch_size=32)
    dataloader_valid = DataLoader(dataset_valid, batch_size=32)

    # define the optimizer
    optimizer = AdamW(model.parameters(), lr=LEARNING_RATE)

    # train the model
    model = train_model(dataloader_train, model, optimizer)

    # evaluate the model on the valid set using compute_metrics_for_classification and print the results
    y_valid_pred = predict(model, dataloader_valid)
    acc = compute_metrics_for_classification(y_valid, y_valid_pred)
    print("final Accuracy on validation set: ", acc)

    # submit predictions for the test set
    submission_df = pd.read_csv('test.csv')
    X_submission = submission_df.Sentence.to_numpy()

    # tokenize and encode sequences for the test set
    encoded_data_submission = tokenizer.batch_encode_plus(
        X_submission.tolist(), 
        add_special_tokens=True, 
        return_attention_mask=False, 
        pad_to_max_length=True, 
        max_length=256, 
        return_tensors='pt'
    )

    # create Tensor dataset
    dataset_submission = TensorDataset(encoded_data_submission['input_ids'])

    # create dataloader
    dataloader_submission = DataLoader(dataset_submission, batch_size=32)

    # make predictions for the test set
    y_submission = predict(model, dataloader_submission)
    submit_predictions_for_test_set(y_submission)
```

[Result]
final Accuracy on validation set:  0.7606837606837606
final Accuracy on test set:  0.7555555555555555

[Experiment Log]
Experiment Summary: In the last step, a transformer-based model using BERT was implemented for the text classification task. The input text was tokenized and passed through the BERT model. The output of the BERT model was then fed into a linear layer for classification. The learning rate and the number of training epochs were tuned to optimize the performance of the model.
Experiment Result: The accuracy of the model on the validation set was 0.7606837606837606.

======Step 3======

[Experiment Plan]
For the next experiment, first, fine-tune the hyperparameters of the model. Specifically, adjust the learning rate and the number of training epochs to see if this improves the model's performance. Second, experiment with a different transformer model. Replace the current 'bert-base-uncased' model with another pre-trained model, such as 'roberta-base', and evaluate its performance on the validation set. Lastly, implement mixed precision training (FP16) to potentially speed up training and reduce memory usage. Ensure to monitor the model's performance closely during these changes.

[Python Script]
```python
import pandas as pd
from sklearn.metrics import accuracy_score
from sklearn.model_selection import train_test_split
from transformers import RobertaTokenizer, RobertaForSequenceClassification, AdamW, get_linear_schedule_with_warmup
from torch.utils.data import DataLoader, TensorDataset
import torch
import numpy as np
import random
from submission import submit_predictions_for_test_set
from torch.cuda.amp import GradScaler, autocast

SEED = 42
LABEL_NUM = 8
LEARNING_RATE = 1e-5
EPOCHS = 6

random.seed(SEED)
np.random.seed(SEED)
torch.manual_seed(SEED)

device = torch.device("cuda" if torch.cuda.is_available() else "cpu")
scaler = GradScaler()

def compute_metrics_for_classification(y_test, y_test_pred):
    acc = accuracy_score(y_test, y_test_pred) 
    return acc

def train_model(dataloader, model, optimizer, scheduler):
    model.train()
    for epoch in range(EPOCHS):
        for batch in dataloader:
            b_input_ids, b_labels = batch[0].to(device), batch[1].to(device)
            model.zero_grad()
            with autocast():
                outputs = model(b_input_ids, labels=b_labels)
                loss = outputs[0]
            scaler.scale(loss).backward()
            scaler.step(optimizer)
            scaler.update()
            scheduler.step()
    return model

def predict(model, dataloader):
    model.eval()
    predictions = []
    for batch in dataloader:
        b_input_ids = batch[0].to(device)
        with torch.no_grad():
            outputs = model(b_input_ids)
        logits = outputs[0]
        logits = logits.detach().cpu().numpy()
        predictions.extend(np.argmax(logits, axis=1).flatten())
    return predictions

if __name__ == '__main__':
    data_df = pd.read_csv('train.csv')
    
    # Process data and store into numpy arrays.
    X = data_df.Sentence.to_numpy()
    y = data_df.Label.to_numpy()

    # Create a train-valid split of the data.
    X_train, X_valid, y_train, y_valid = train_test_split(X, y, test_size=0.20, random_state=SEED)

    # define the tokenizer and model
    tokenizer = RobertaTokenizer.from_pretrained('roberta-base', do_lower_case=True)
    model = RobertaForSequenceClassification.from_pretrained('roberta-base', num_labels=LABEL_NUM)
    model.to(device)

    # tokenize and encode sequences for the training and validation sets
    encoded_data_train = tokenizer.batch_encode_plus(
        X_train.tolist(), 
        add_special_tokens=True, 
        return_attention_mask=False, 
        pad_to_max_length=True, 
        max_length=256, 
        return_tensors='pt'
    )

    encoded_data_valid = tokenizer.batch_encode_plus(
        X_valid.tolist(), 
        add_special_tokens=True, 
        return_attention_mask=False, 
        pad_to_max_length=True, 
        max_length=256, 
        return_tensors='pt'
    )

    # create Tensor datasets
    dataset_train = TensorDataset(encoded_data_train['input_ids'], torch.tensor(y_train))
    dataset_valid = TensorDataset(encoded_data_valid['input_ids'], torch.tensor(y_valid))

    # create dataloaders
    dataloader_train = DataLoader(dataset_train, batch_size=32)
    dataloader_valid = DataLoader(dataset_valid, batch_size=32)

    # define the optimizer
    optimizer = AdamW(model.parameters(), lr=LEARNING_RATE)

    # create the learning rate scheduler
    scheduler = get_linear_schedule_with_warmup(optimizer, num_warmup_steps=0, num_training_steps=len(dataloader_train)*EPOCHS)

    # train the model
    model = train_model(dataloader_train, model, optimizer, scheduler)

    # evaluate the model on the valid set using compute_metrics_for_classification and print the results
    y_valid_pred = predict(model, dataloader_valid)
    acc = compute_metrics_for_classification(y_valid, y_valid_pred)
    print("final Accuracy on validation set: ", acc)

    # submit predictions for the test set
    submission_df = pd.read_csv('test.csv')
    X_submission = submission_df.Sentence.to_numpy()

    # tokenize and encode sequences for the test set
    encoded_data_submission = tokenizer.batch_encode_plus(
        X_submission.tolist(), 
        add_special_tokens=True, 
        return_attention_mask=False, 
        pad_to_max_length=True, 
        max_length=256, 
        return_tensors='pt'
    )

    # create Tensor dataset
    dataset_submission = TensorDataset(encoded_data_submission['input_ids'])

    # create dataloader
    dataloader_submission = DataLoader(dataset_submission, batch_size=32)

    # make predictions for the test set
    y_submission = predict(model, dataloader_submission)
    submit_predictions_for_test_set(y_submission)
```

[Result]
final Accuracy on validation set:  0.47863247863247865
final Accuracy on test set:  0.48376068376068376

[Experiment Log]
Experiment Summary: In the last step, the hyperparameters of the model were fine-tuned by adjusting the learning rate and the number of training epochs. Additionally, a different transformer model, 'roberta-base', was used instead of the 'bert-base-uncased' model. Mixed precision training (FP16) was also implemented to potentially speed up training and reduce memory usage.

Experiment Result: The accuracy of the model on the validation set after these changes was 0.47863247863247865.

======Step 4======
[Experiment Plan]
In the next experiment, adjust the learning rate to a higher value, for example, 1e-4, and increase the number of training epochs to a larger number, for instance, 10. Keep all other parameters and settings the same. After training, evaluate the model on the validation set and compare the accuracy with the previous results to assess the effect of these changes.

[Python Script]
```python
import pandas as pd
from sklearn.metrics import accuracy_score
from sklearn.model_selection import train_test_split
from transformers import RobertaTokenizer, RobertaForSequenceClassification, AdamW, get_linear_schedule_with_warmup
from torch.utils.data import DataLoader, TensorDataset
import torch
import numpy as np
import random
from submission import submit_predictions_for_test_set
from torch.cuda.amp import GradScaler, autocast

SEED = 42
LABEL_NUM = 8
LEARNING_RATE = 1e-4  # Adjusted learning rate
EPOCHS = 10  # Increased number of epochs

random.seed(SEED)
np.random.seed(SEED)
torch.manual_seed(SEED)

device = torch.device("cuda" if torch.cuda.is_available() else "cpu")
scaler = GradScaler()

def compute_metrics_for_classification(y_test, y_test_pred):
    acc = accuracy_score(y_test, y_test_pred) 
    return acc

def train_model(dataloader, model, optimizer, scheduler):
    model.train()
    for epoch in range(EPOCHS):
        for batch in dataloader:
            b_input_ids, b_labels = batch[0].to(device), batch[1].to(device)
            model.zero_grad()
            with autocast():
                outputs = model(b_input_ids, labels=b_labels)
                loss = outputs[0]
            scaler.scale(loss).backward()
            scaler.step(optimizer)
            scaler.update()
            scheduler.step()
    return model

def predict(model, dataloader):
    model.eval()
    predictions = []
    for batch in dataloader:
        b_input_ids = batch[0].to(device)
        with torch.no_grad():
            outputs = model(b_input_ids)
        logits = outputs[0]
        logits = logits.detach().cpu().numpy()
        predictions.extend(np.argmax(logits, axis=1).flatten())
    return predictions

if __name__ == '__main__':
    data_df = pd.read_csv('train.csv')
    
    # Process data and store into numpy arrays.
    X = data_df.Sentence.to_numpy()
    y = data_df.Label.to_numpy()

    # Create a train-valid split of the data.
    X_train, X_valid, y_train, y_valid = train_test_split(X, y, test_size=0.20, random_state=SEED)

    # define the tokenizer and model
    tokenizer = RobertaTokenizer.from_pretrained('roberta-base', do_lower_case=True)
    model = RobertaForSequenceClassification.from_pretrained('roberta-base', num_labels=LABEL_NUM)
    model.to(device)

    # tokenize and encode sequences for the training and validation sets
    encoded_data_train = tokenizer.batch_encode_plus(
        X_train.tolist(), 
        add_special_tokens=True, 
        return_attention_mask=False, 
        pad_to_max_length=True, 
        max_length=256, 
        return_tensors='pt'
    )

    encoded_data_valid = tokenizer.batch_encode_plus(
        X_valid.tolist(), 
        add_special_tokens=True, 
        return_attention_mask=False, 
        pad_to_max_length=True, 
        max_length=256, 
        return_tensors='pt'
    )

    # create Tensor datasets
    dataset_train = TensorDataset(encoded_data_train['input_ids'], torch.tensor(y_train))
    dataset_valid = TensorDataset(encoded_data_valid['input_ids'], torch.tensor(y_valid))

    # create dataloaders
    dataloader_train = DataLoader(dataset_train, batch_size=32)
    dataloader_valid = DataLoader(dataset_valid, batch_size=32)

    # define the optimizer
    optimizer = AdamW(model.parameters(), lr=LEARNING_RATE)

    # create the learning rate scheduler
    scheduler = get_linear_schedule_with_warmup(optimizer, num_warmup_steps=0, num_training_steps=len(dataloader_train)*EPOCHS)

    # train the model
    model = train_model(dataloader_train, model, optimizer, scheduler)

    # evaluate the model on the valid set using compute_metrics_for_classification and print the results
    y_valid_pred = predict(model, dataloader_valid)
    acc = compute_metrics_for_classification(y_valid, y_valid_pred)
    print("final Accuracy on validation set: ", acc)

    # submit predictions for the test set
    submission_df = pd.read_csv('test.csv')
    X_submission = submission_df.Sentence.to_numpy()

    # tokenize and encode sequences for the test set
    encoded_data_submission = tokenizer.batch_encode_plus(
        X_submission.tolist(), 
        add_special_tokens=True, 
        return_attention_mask=False, 
        pad_to_max_length=True, 
        max_length=256, 
        return_tensors='pt'
    )

    # create Tensor dataset
    dataset_submission = TensorDataset(encoded_data_submission['input_ids'])

    # create dataloader
    dataloader_submission = DataLoader(dataset_submission, batch_size=32)

    # make predictions for the test set
    y_submission = predict(model, dataloader_submission)
    submit_predictions_for_test_set(y_submission)
```

[Result]
final Accuracy on validation set:  0.905982905982906
final Accuracy on test set:  0.9384615384615385

[Experiment Log]
Experiment Summary: In the last step, a simple baseline model for text classification was defined using logistic regression. The `train_model` function was modified to include a `CountVectorizer` to convert the text into a matrix of token counts. The model was trained using the training data and evaluated on the validation data.
Experiment Result: The accuracy of the model on the validation data was 0.7521367521367521.

======Step 5======
[Experiment Plan]
For the next experiment trial, replace the 'roberta-base' transformer model with the 'deberta-v3-large' model in the current Python code. Also, conduct a hyperparameter tuning for the learning rate. Start with a smaller learning rate (e.g., 1e-5) and monitor the model's performance. If the accuracy improves, continue with this learning rate; otherwise, try a slightly larger learning rate (e.g., 3e-5). Remember to keep all other parameters constant while tuning the learning rate to isolate its effects on the model's performance.

[Python Script]
```python
import pandas as pd
from sklearn.metrics import accuracy_score
from sklearn.model_selection import train_test_split
from transformers import DebertaTokenizer, DebertaForSequenceClassification, get_linear_schedule_with_warmup
from torch.utils.data import DataLoader, TensorDataset
import torch
import numpy as np
import random
from submission import submit_predictions_for_test_set
from torch.cuda.amp import GradScaler, autocast

SEED = 42
LABEL_NUM = 8
LEARNING_RATE = 1e-5  # Adjusted learning rate
EPOCHS = 10  # Increased number of epochs
BATCH_SIZE = 16  # Reduced batch size

random.seed(SEED)
np.random.seed(SEED)
torch.manual_seed(SEED)

device = torch.device("cuda" if torch.cuda.is_available() else "cpu")
scaler = GradScaler()

def compute_metrics_for_classification(y_test, y_test_pred):
    acc = accuracy_score(y_test, y_test_pred) 
    return acc

def train_model(dataloader, model, optimizer, scheduler):
    model.train()
    for epoch in range(EPOCHS):
        for batch in dataloader:
            b_input_ids, b_labels = batch[0].to(device), batch[1].to(device)
            model.zero_grad()
            with autocast():
                outputs = model(b_input_ids, labels=b_labels)
                loss = outputs[0]
            scaler.scale(loss).backward()
            scaler.step(optimizer)
            scaler.update()
            scheduler.step()
    return model

def predict(model, dataloader):
    model.eval()
    predictions = []
    for batch in dataloader:
        b_input_ids = batch[0].to(device)
        with torch.no_grad():
            outputs = model(b_input_ids)
        logits = outputs[0]
        logits = logits.detach().cpu().numpy()
        predictions.extend(np.argmax(logits, axis=1).flatten())
    return predictions

if __name__ == '__main__':
    data_df = pd.read_csv('train.csv')
    
    # Process data and store into numpy arrays.
    X = data_df.Sentence.to_numpy()
    y = data_df.Label.to_numpy()

    # Create a train-valid split of the data.
    X_train, X_valid, y_train, y_valid = train_test_split(X, y, test_size=0.20, random_state=SEED)

    # define the tokenizer and model
    tokenizer = DebertaTokenizer.from_pretrained('microsoft/deberta-large', do_lower_case=True)
    model = DebertaForSequenceClassification.from_pretrained('microsoft/deberta-large', num_labels=LABEL_NUM)
    model.to(device)

    # tokenize and encode sequences for the training and validation sets
    encoded_data_train = tokenizer.batch_encode_plus(
        X_train.tolist(), 
        add_special_tokens=True, 
        return_attention_mask=True, 
        padding=True, 
        truncation=True,
        max_length=256, 
        return_tensors='pt'
    )

    encoded_data_valid = tokenizer.batch_encode_plus(
        X_valid.tolist(), 
        add_special_tokens=True, 
        return_attention_mask=True, 
        padding=True, 
        truncation=True,
        max_length=256, 
        return_tensors='pt'
    )

    # create Tensor datasets
    dataset_train = TensorDataset(encoded_data_train['input_ids'], torch.tensor(y_train))
    dataset_valid = TensorDataset(encoded_data_valid['input_ids'], torch.tensor(y_valid))

    # create dataloaders
    dataloader_train = DataLoader(dataset_train, batch_size=BATCH_SIZE)
    dataloader_valid = DataLoader(dataset_valid, batch_size=BATCH_SIZE)

    # define the optimizer
    optimizer = torch.optim.AdamW(model.parameters(), lr=LEARNING_RATE)

    # create the learning rate scheduler
    scheduler = get_linear_schedule_with_warmup(optimizer, num_warmup_steps=0, num_training_steps=len(dataloader_train)*EPOCHS)

    # train the model
    model = train_model(dataloader_train, model, optimizer, scheduler)

    # evaluate the model on the valid set using compute_metrics_for_classification and print the results
    y_valid_pred = predict(model, dataloader_valid)
    acc = compute_metrics_for_classification(y_valid, y_valid_pred)
    print("final Accuracy on validation set: ", acc)

    # submit predictions for the test set
    submission_df = pd.read_csv('test.csv')
    X_submission = submission_df.Sentence.to_numpy()

    # tokenize and encode sequences for the test set
    encoded_data_submission = tokenizer.batch_encode_plus(
        X_submission.tolist(), 
        add_special_tokens=True, 
        return_attention_mask=True, 
        padding=True, 
        truncation=True,
        max_length=256, 
        return_tensors='pt'
    )

    # create Tensor dataset
    dataset_submission = TensorDataset(encoded_data_submission['input_ids'])

    # create dataloader
    dataloader_submission = DataLoader(dataset_submission, batch_size=BATCH_SIZE)

    # make predictions for the test set
    y_submission = predict(model, dataloader_submission)
    submit_predictions_for_test_set(y_submission)
```

[Result]
final Accuracy on validation set:  0.9487179487179487
final Accuracy on test set:  0.9846153846153847

[Experiment Log]
Experiment Summary: In the last step, the 'roberta-base' transformer model was replaced with the 'deberta-v3-large' model in the Python code. Additionally, hyperparameter tuning was conducted for the learning rate. The learning rate was initially set to 1e-5 and the model's performance was monitored. 
Experiment Result: The accuracy of the model on the validation set after these changes was 0.9487179487179487.
\end{lstlisting}

\subsubsection{Case Study in Deployment Stage}
Then, we turn to the case study of DS-Agent in the deployment stage, which solves the electricity task by making minor modifications to the case solution from a similar task, ett-m2. As shown in the case below, the solution case collected from the development stage solves a time series forecasting task by utilizing the ensemble model of Bi-GRU and Bi-LSTM. Given a similar time series forecasting task, DS-Agent in the deployment stage only needs to make minor modifications to the original Python script for adaptation, thereby significantly reducing the foundational capabilities of LLMs.

\begin{lstlisting}[breaklines,basicstyle=\ttfamily\scriptsize,frame=single,]

[Solution Case]

Task Description:
You are solving this machine learning tasks of time series forecasting: 
The dataset presented here (the ETTm2 dataset) comprises real-world time series data. We have splitted the dataset into three parts of train, valid and test. The input is a sequence of past observation with fixed size (INPUT_SEQ_LEN=96, INPUT_DIM=7). Your task is to predict the next future sequence with fixed size (PRED_SEQ_LEN=96, PRED_DIM=7). The evaluation metric is mean squred loss (MSE) and mean absolute error (MAE).
We provide an overall pipeline in train.py. Now fill in the provided train.py script to train a time series forecasting model to get a good performance on the given fixed sequences.

Python Script:
```python
import torch
import numpy as np
import random
from torch import nn, optim
from torch.utils.data import TensorDataset, DataLoader
from submission import submit_predictions_for_test_set
from dataset import get_dataset
from torch.cuda.amp import autocast, GradScaler

SEED = 42
random.seed(SEED)
torch.manual_seed(SEED)
np.random.seed(SEED)

INPUT_SEQ_LEN = 96
INPUT_DIM = 7
PRED_SEQ_LEN = 96
PRED_DIM = 7
HIDDEN_DIM = 32
NUM_LAYERS = 3
BATCH_SIZE = 64
EPOCHS = 10

device = torch.device("cuda" if torch.cuda.is_available() else "cpu")

def compute_metrics_for_time_series_forecasting(y_test, y_test_pred):
    y_test = y_test.reshape(-1, PRED_SEQ_LEN, PRED_DIM)
    y_test_pred = y_test_pred.reshape(-1, PRED_SEQ_LEN, PRED_DIM)
    mae = np.mean(np.abs(y_test - y_test_pred))
    mse = np.mean((y_test - y_test_pred)**2)
    return mse, mae

class BiGRU(nn.Module):
    def __init__(self, input_dim, hidden_dim, num_layers, output_dim):
        super(BiGRU, self).__init__()
        self.hidden_dim = hidden_dim
        self.num_layers = num_layers

        self.gru = nn.GRU(input_dim, hidden_dim, num_layers, batch_first=True, bidirectional=True)
        self.fc = nn.Linear(hidden_dim * 2, output_dim) # 2 for bidirection

    def forward(self, x):
        h0 = torch.zeros(self.num_layers * 2, x.size(0), self.hidden_dim).to(device) # 2 for bidirection
        out, _ = self.gru(x, h0)
        out = self.fc(out)
        return out

class BiLSTM(nn.Module):
    def __init__(self, input_dim, hidden_dim, num_layers, output_dim):
        super(BiLSTM, self).__init__()
        self.hidden_dim = hidden_dim
        self.num_layers = num_layers

        self.lstm = nn.LSTM(input_dim, hidden_dim, num_layers, batch_first=True, bidirectional=True)
        self.fc = nn.Linear(hidden_dim * 2, output_dim) # 2 for bidirection

    def forward(self, x):
        h0 = torch.zeros(self.num_layers * 2, x.size(0), self.hidden_dim).to(device) # 2 for bidirection
        c0 = torch.zeros(self.num_layers * 2, x.size(0), self.hidden_dim).to(device) # 2 for bidirection
        out, _ = self.lstm(x, (h0, c0))
        out = self.fc(out)
        return out


def train_model(model, X_train, y_train, X_valid, y_valid):
    criterion = nn.L1Loss() # Change loss function to Mean Absolute Error (MAE)
    optimizer = optim.Adam(model.parameters(), lr=0.001)
    scaler = GradScaler()
    scheduler = torch.optim.lr_scheduler.StepLR(optimizer, step_size=1, gamma=0.1)
    
    train_data = TensorDataset(torch.tensor(X_train).float(), torch.tensor(y_train).float())
    train_loader = DataLoader(train_data, batch_size=BATCH_SIZE, shuffle=True)
    
    valid_data = TensorDataset(torch.tensor(X_valid).float(), torch.tensor(y_valid).float())
    valid_loader = DataLoader(valid_data, batch_size=BATCH_SIZE)

    for epoch in range(EPOCHS):
        model.train()
        for X, y in train_loader:
            X, y = X.to(device), y.to(device)
            optimizer.zero_grad()
            
            with autocast():
                output = model(X)
                loss = criterion(output, y)
            
            scaler.scale(loss).backward()
            scaler.step(optimizer)
            scaler.update()

        scheduler.step()

        model.eval()
        with torch.no_grad():
            valid_losses = []
            mses = []
            maes = []
            for X, y in valid_loader:
                X, y = X.to(device), y.to(device)
                valid_output = model(X)
                valid_loss = criterion(valid_output, y)
                valid_losses.append(valid_loss.item())
                mse, mae = compute_metrics_for_time_series_forecasting(y.cpu().numpy(), valid_output.cpu().numpy())
                mses.append(mse)
                maes.append(mae)
            print(f"Epoch {epoch+1}, Train Loss: {loss.item()}, Valid Loss: {np.mean(valid_losses)}, MSE: {np.mean(mses)}, MAE: {np.mean(maes)}")

    return model, np.mean(valid_losses)

def predict(model, X):
    model.eval()
    X = torch.tensor(X).float().to(device)
    with torch.no_grad():
        preds = model(X)
    return preds.cpu().numpy()

if __name__ == '__main__':
    # Load training set
    X_train, y_train = get_dataset(flag='train')
    # Load validation set
    X_valid, y_valid = get_dataset(flag='val')

    # define and train the GRU model
    gru_model = BiGRU(INPUT_DIM, HIDDEN_DIM, NUM_LAYERS, PRED_DIM).to(device)
    gru_model, gru_valid_loss = train_model(gru_model, X_train, y_train, X_valid, y_valid)

    # define and train the LSTM model
    lstm_model = BiLSTM(INPUT_DIM, HIDDEN_DIM, NUM_LAYERS, PRED_DIM).to(device)
    lstm_model, lstm_valid_loss = train_model(lstm_model, X_train, y_train, X_valid, y_valid)

    # combine the predictions of the GRU and LSTM models
    y_valid_pred_gru = predict(gru_model, X_valid)
    y_valid_pred_lstm = predict(lstm_model, X_valid)
    gru_weight = 1 / gru_valid_loss
    lstm_weight = 1 / lstm_valid_loss
    total_weight = gru_weight + lstm_weight
    y_valid_pred = (y_valid_pred_gru * gru_weight + y_valid_pred_lstm * lstm_weight) / total_weight

    # evaluate the performance of this ensemble method on the validation set
    mse, mae = compute_metrics_for_time_series_forecasting(y_valid, y_valid_pred)
    print(f"Final MSE on validation set: {mse}, Final MAE on validation set: {mae}.")

    # Submit predictions on the test set
    X_test, y_test = get_dataset(flag='test')
    y_test_pred_gru = predict(gru_model, X_test)
    y_test_pred_lstm = predict(lstm_model, X_test)
    y_test_pred = (y_test_pred_gru * gru_weight + y_test_pred_lstm * lstm_weight) / total_weight
    submit_predictions_for_test_set(y_test, y_test_pred)
```

[Deployment Task]

Task Description:
You are solving this machine learning tasks of time series forecasting: 
The dataset presented here (the Electricity dataset) comprises real-world time series data. We have splitted the dataset into three parts of train, valid and test. The input is a sequence of past observation with fixed size (INPUT_SEQ_LEN=96, INPUT_DIM=321). Your task is to predict the next future sequence with fixed size (PRED_SEQ_LEN=96, PRED_DIM=321). The evaluation metric is mean squred loss (MSE) and mean absolute error (MAE).
We provide an overall pipeline in train.py. Now fill in the provided train.py script to train a time series forecasting model to get a good performance on the given fixed sequences.

Python Script:
```python

import torch
import numpy as np
import random
from torch import nn, optim
from torch.utils.data import TensorDataset, DataLoader
from submission import submit_predictions_for_test_set
from dataset import get_dataset
from torch.cuda.amp import autocast, GradScaler

SEED = 42
random.seed(SEED)
torch.manual_seed(SEED)
np.random.seed(SEED)

INPUT_SEQ_LEN = 96
INPUT_DIM = 321
PRED_SEQ_LEN = 96
PRED_DIM = 321
HIDDEN_DIM = 64
NUM_LAYERS = 3
BATCH_SIZE = 64
EPOCHS = 10

device = torch.device("cuda" if torch.cuda.is_available() else "cpu")

def compute_metrics_for_time_series_forecasting(y_test, y_test_pred):
    y_test = y_test.reshape(-1, PRED_SEQ_LEN, PRED_DIM)
    y_test_pred = y_test_pred.reshape(-1, PRED_SEQ_LEN, PRED_DIM)
    mae = np.mean(np.abs(y_test - y_test_pred))
    mse = np.mean((y_test - y_test_pred)**2)
    return mae, mse

class BiGRU(nn.Module):
    def __init__(self, input_dim, hidden_dim, num_layers, output_dim):
        super(BiGRU, self).__init__()
        self.hidden_dim = hidden_dim
        self.num_layers = num_layers

        self.gru = nn.GRU(input_dim, hidden_dim, num_layers, batch_first=True, bidirectional=True)
        self.fc = nn.Linear(hidden_dim * 2, output_dim) # 2 for bidirection

    def forward(self, x):
        h0 = torch.zeros(self.num_layers * 2, x.size(0), self.hidden_dim).to(device) # 2 for bidirection
        out, _ = self.gru(x, h0)
        out = self.fc(out)
        return out

def train_model(model, X_train, y_train, X_valid, y_valid):
    criterion = nn.L1Loss() # Change loss function to Mean Absolute Error (MAE)
    optimizer = optim.Adam(model.parameters(), lr=0.001)
    scaler = GradScaler()
    scheduler = torch.optim.lr_scheduler.StepLR(optimizer, step_size=1, gamma=0.1)
    
    train_data = TensorDataset(torch.tensor(X_train).float(), torch.tensor(y_train).float())
    train_loader = DataLoader(train_data, batch_size=BATCH_SIZE, shuffle=True)
    
    valid_data = TensorDataset(torch.tensor(X_valid).float(), torch.tensor(y_valid).float())
    valid_loader = DataLoader(valid_data, batch_size=BATCH_SIZE)

    for epoch in range(EPOCHS):
        model.train()
        for X, y in train_loader:
            X, y = X.to(device), y.to(device)
            optimizer.zero_grad()
            
            with autocast():
                output = model(X)
                loss = criterion(output, y)
            
            scaler.scale(loss).backward()
            scaler.step(optimizer)
            scaler.update()

        scheduler.step()

        model.eval()
        with torch.no_grad():
            valid_losses = []
            mses = []
            maes = []
            for X, y in valid_loader:
                X, y = X.to(device), y.to(device)
                valid_output = model(X)
                valid_loss = criterion(valid_output, y)
                valid_losses.append(valid_loss.item())
                mae, mse = compute_metrics_for_time_series_forecasting(y.cpu().numpy(), valid_output.cpu().numpy())
                mses.append(mse)
                maes.append(mae)
            print(f"Epoch {epoch+1}, Train Loss: {loss.item()}, Valid Loss: {np.mean(valid_losses)}, MSE: {np.mean(mses)}, MAE: {np.mean(maes)}")

    return model

def predict(model, X):
    model.eval()
    X = torch.tensor(X).float().to(device)
    with torch.no_grad():
        preds = model(X)
    return preds.cpu().numpy()

if __name__ == '__main__':
    # Load training set
    X_train, y_train = get_dataset(flag='train')
    # Load validation set
    X_valid, y_valid = get_dataset(flag='val')

    # define and train the model
    model = BiGRU(INPUT_DIM, HIDDEN_DIM, NUM_LAYERS, PRED_DIM).to(device)
    model = train_model(model, X_train, y_train, X_valid, y_valid)

    # evaluate the model on the valid set using compute_metrics_for_time_series_forecasting and print the results
    y_valid_pred = predict(model, X_valid)
    mae, mse = compute_metrics_for_time_series_forecasting(y_valid, y_valid_pred)
    print(f"Final MSE on validation set: {mse}, Final MAE on validation set: {mae}.")

    # Submit predictions on the test set
    X_test, y_test = get_dataset(flag='test')
    y_test_pred = predict(model, X_test)
    submit_predictions_for_test_set(y_test, y_test_pred)
```

\end{lstlisting}

\subsection{Error Mode Analyses}
\label{app:ema}
In this section, we analyze the traces of DS-Agent and present detailed failure mode analyses for both development stage and deployment stage. Besides, we also present some bad cases at \url{https://github.com/guosyjlu/DS-Agent/blob/main/failure-case.md}

\subsubsection{Error Mode Analyses in Development Stage}
For the development stage, DS-Agent with GPT-4 can achieve 100\% success rate due to the strong foundational capabilities of GPT-4, while DS-Agent with GPT-3.5 may fail due to these failure modes:
\begin{itemize}
    \item \textbf{Unreasonable plans}, where DS-Agent generates unreasonable plans for the task.
    \item \textbf{Overly complex plans}, where DS-Agent generates too complicated plans that Programmer and Debugger cannot accomplish.
    \item \textbf{Failure in debugging}, where Debugger cannot successfully rectify errors in the script.
\end{itemize}

\begin{table}[h]
\caption{Error mode statistics of DS-Agent in the development stage.}
\label{tab:ema-development}
\centering
\begin{tabular}{@{}lccc@{}}
\toprule
\textbf{}          & \textbf{Unreasonable Plans} & \textbf{Overly Complex Plans} & \textbf{Failure in Debugging} \\ \midrule
DS-Agent (GPT-3.5) & 5                           & 2                             & 8                             \\
DS-Agent (GPT-4)   & 0                           & 0                             & 0                             \\ \bottomrule
\end{tabular}
\end{table}

As shown in Table \ref{tab:ema-development}, DS-Agent with GPT-3.5 tends to fail due to its relatively weaker foundational capabilities. This issue can be mitigated by using stronger LLMs or incorporating further finetuning for better alignment.

\subsubsection{Error Mode Analyses in Deployment Stage}

For the deployment stage, DS-Agent with GPT-4, GPT-3.5, and Mixtral-8x7b-Instruct present the following failure modes:

\begin{itemize}
    \item \textbf{Shape mismatch}, where DS-Agent fails to correctly align shapes within the neural networks.
    \item \textbf{Undefined variables}, where DS-Agent attempts to use variables that have not been defined within the script.
    \item \textbf{Incorrect function calling}, where DS-Agent incorrectly calls functions with parameters that are not passed properly.
    \item \textbf{Missing package import}, where DS-Agent attempts to utilize a package that has not been imported.
    \item \textbf{Key error}, where DS-Agent attempts to access keys in the dataframe that do not exist.
    \item \textbf{Data type misalignment}, where DS-Agent performs calculations on tensors with incompatible data types.
    \item \textbf{Program incompleteness}, where DS-Agent is unable to generate complete programs.
\end{itemize}

\begin{table}[h]
\caption{Error mode statistics of DS-Agent in the deployment stage.}
\label{tab:ema-deployment}
\centering
\begin{tabular}{@{}lccc@{}}
\toprule
\textbf{DS-Agent}          & \textbf{Mixtral-8x7b-Instruct} & \textbf{GPT-3.5} & \textbf{GPT-4} \\ \midrule
Shape Mismatch             & 36                             & 17               & 1              \\
Undefined Variables        & 7                              & 1                & 0              \\
Incorrect Function Calling & 6                              & 3                & 0              \\
Missing Package Import     & 43                             & 5                & 0              \\
Key Error                  & 4                              & 0                & 0              \\
Data Type Misalignment     & 1                              & 0                & 0              \\
Program Incompleteness     & 27                             & 0                & 0              \\ \bottomrule
\end{tabular}
\end{table}

As shown in Table \ref{tab:ema-deployment}, DS-Agent may suffer from bugs with only a single trial in the deployment stage. This issue can be solved by using stronger LLMs, incorporating extra debugging process, etc.

\subsection{Some Discussions on DS-Agent}
\noindent\textbf{Would it be an issue for the case bank to have a considerable fraction of LLM-generated solutions leading to an echo chamber amplifying LLM biases?} There indeed exists a potential risk of amplifying LLM biases when LLM agents reuse LLM-generated solutions to solve tasks. However, this echo chamber can be broken from three perspectives: (1) Since DS-Agent only collects solutions with high performance, the likelihood of collecting low-quality or biased solutions is reduced. Thus, this selective process acts as a quality filter, ensuring that only the most effective solutions are collected. (2) DS-Agent benefits from the CBR loop to iteratively explore and revise the model design based on the execution feedback. This process encourages the discovery of novel combinations of ML techniques, which can help to break the cycle of amplifying existing biases and promote a more diverse set of solutions. (3) LLMs tend to use past ML techniques to organize the solutions. However, since there are so many cutting-edged techniques emerging every day, humans may always propose more innovative solutions that achieve stronger empirical results. As such, LLM-generated solutions cannot considerably cover the leaderboard, and thus, human solutions can be collected.

\noindent\textbf{Dependency on external sources.} When external resources are unavailable, DS-Agent can handle this setting via two possible ways: (1) Relying on the intrinsic knowledge of LLMs, which is precisely the ablation variant in Section 4.2.2, DS-Agent w/o CBR. However, the performance of this strategy fully relies on the foundational capabilities of LLMs. (2) Human-AI corporation, where human experts provide the experiment plans and DS-Agent is responsible for implementation. However, this strategy may rely on human experts.

\section{Prompt for DS-Agent}
\label{app:prompt}
In this section, we provide the prompt design for each step in DS-Agent, including \texttt{RankReviser}, \texttt{Planner}, \texttt{Programmer}, \texttt{Debugger}, \texttt{Logger} in the development stage, and \texttt{Adapter} in the deployment stage.

\subsection{RankReviser}
\begin{tcolorbox}[title={\textbf{\small RankReviser}}, 
boxrule=2pt, arc=0mm]
{\scriptsize
\begin{lstlisting}[breaklines,basicstyle=\ttfamily,frame=single,]
You are a helpful intelligent system that can identify the informativeness of some cases given a data science problem and experiment log.
Task Description: ```
{Here is the task description}
```
Experiment Log: ```
{Here is the experiment log}
```
Here are some solution cases relevant to this research problem, each indicated by number identifier [].
[1] ```
{Here is the first case}
```
[2] ```
{Here is the second case}
```
[3] ```
{Here is the third case}
```
[4] ```
{Here is the fourth case}
```
[5]```
{Here is the fifth case}
```
Rank 5 cases above based on their relevance, informativess and helpfulness to the task description and the experiment log for planning the next experiment step. The cases should be listed in descending order using identifiers. The most relevant, informative and helpful case should be listed first. The output format should be [] > [], e.g., [1] > [2]. Only response the ranking results, do not say any word or explain.
\end{lstlisting}
}
\end{tcolorbox}

\subsection{Planner}
\begin{tcolorbox}[title={\textbf{\small Planner}}, 
boxrule=2pt, arc=0mm]
{\scriptsize
\begin{lstlisting}[breaklines,basicstyle=\ttfamily,frame=single,]
You are a helpful AI expert assistant, responsible for decision making on the experiment plans. You have the following information including, task description, experiment log, python code, and a relevant case so far. 
The task description is: 
``` Task Description:
{Here is the task description}
```
The current experiment log is:
``` Current experiment log:
{Here is the experiment log}
```
The python code of last step experiment for the current task description is:
```python
{Here is the Python script}
```
Here is a past experience case written by an human expert for a relevant (but not the same) task:
``` Case:
{Here is the retrieved case}
```
Follow these instructions and do not forget them:
- Incrementally introduce new techniques in your plans to solve the task description, since the programmer who follows your decision cannot handle too many instructions at one time.
- Do not include any technique or step in [Decision] that have been implemented as revealed in the python code.
- Focus on decision making of next single step of experiment. Do not include plans in [Decision] that requires mutiple experiment trials.
- Make sure [Decision] includes all the key points for next step experiment.
- Highlight the supporting experiment results and reasoning before drawing any conclusions.
Make sure that the following prohibitions are not violated:
- Never perform any visualization analysis, since you do not have the ability to view the figures. 
- Never change the way of the dataset split in any way during the experiment.
- Never introduce any new features, unless you have enough knowledge of the features and their meanings.
- Never tune more than two hyper-parameters in one experiment step, since this will lead to computation costs.
- Never introduce any technique for distributed training. We only have one single GPU card.

Please carefully reason over this relevant case and the provided task description, and then response exactly in the following format:
[Reflection]: What is the progress of the experiment for this task description? What does the current experiment log and python code reveal?
[Reasoning]: How can the current task description benefit from the relevant case?
[Thought]: To solve this task description and iteratively improve the performance, what is the plans for next experiment trial?
[Check]: List all plans in [Thought] and carefully check (1) whether the plan needs multiple experiment trials; (2) has been implemented in the current python code; or (3) violates the listed prohibitions above.
[Decision]: Give a short, precise but detailed instruction summary on the final experiment plan in next single trial.
  
\end{lstlisting}
}
\end{tcolorbox}

\subsection{Programmer}
\begin{tcolorbox}[title={\textbf{\small Programmer}}, 
boxrule=2pt, arc=0mm]
{\scriptsize
\begin{lstlisting}[breaklines,basicstyle=\ttfamily,frame=single,]
You are a helpful AI-oriented programming expert. Now, we are solving a data science task. Given this python script:
```python 
 {Here is the Python script}
```
Now please edit this script according to the following instructions:
```instruction
{Here is the experiment plan}
```
Note that you should provide the **full** code after the edit, making no other changes. Please ensure the completeness of the codes so that it can be run without additional modifications. Your codes will be executed with the support of a NVIDIA GPU card with 24 GB memory. 
Please response exactly in the following format:
```python
Here is the python code.
``` 
\end{lstlisting}
}
\end{tcolorbox}

\subsection{Debugger}
\begin{tcolorbox}[title={\textbf{\small Debugger}}, 
boxrule=2pt, arc=0mm]
{\scriptsize
\begin{lstlisting}[breaklines,basicstyle=\ttfamily,frame=single,]
You are a helpful AI-oriented programming expert. Now, we are solving a data science task. Given this original python script:
```python 
{Here is the original Python script}
```
The instruction for modification is:
```instruction
{Here is the experiment plan}
```
This is the current python code:
```python
{Here is the Python script that has bugs}
````
However, there are some bugs in this version. Here is the execution log:
```log
{Here is the output result}
```
Please revise the script to fix these bugs. Note that you should provide the **full** code after the edit, making no other changes. Please ensure the completeness of the codes so that it can be run without additional modifications. Your codes will be executed with the support of a NVIDIA GPU card with 24 GB memory. 
Please response exactly in the following format:
```reflection
What leads to error or exception on the last modification version? How to fix it?
```
```python
Provide the corrected python code here.
```   
\end{lstlisting}
}
\end{tcolorbox}

\subsection{Logger}
\begin{tcolorbox}[title={\textbf{\small Logger}}, 
boxrule=2pt, arc=0mm]
{\scriptsize
\begin{lstlisting}[breaklines,basicstyle=\ttfamily,frame=single,]
Given instructions (what is expected to do), execution log (the experimental results) and the code difference (what is actually done and this will be nothing if the experiment failed) of last experiment on the task: 
{Here is the experiment plan} 
[Execution Log]:
```
{Here is the output result}
```
[Code Difference]:
```
{Here is the code difference between the codes of this step and last step}
```
Here is the running log of your experiment:
[Running Log]:
```
{Here is the experiment log of the last step}
```
Summarize and append the progress of the last step to the running log in this format:
[Experiment Summary]: According to the instructions and the code difference, summarize what was experimented in the last step objectively.
[Experiment Result]: According to the execution log and the running log, summarize if the last step of experiment brings performance improvement objectively. Only report the performance if this is the first experiment result.
Do not include any result that is guessed rather than directly confirmed by the observation. Do not include additional information or suggestions.  
\end{lstlisting}
}
\end{tcolorbox}

\subsection{Adapter}
\begin{tcolorbox}[title={\textbf{\small Adapter}}, 
boxrule=2pt, arc=0mm]
{\scriptsize
\begin{lstlisting}[breaklines,basicstyle=\ttfamily,frame=single,]
Here are some example cases that solve data science tasks:
[Task]
{Here is the task description of the case}
[train.py] ```python
{Here is the original Python script of the case}
```
[Solution] ```python
{Here is the solution code of the case}
``` 
Now please solve the following data science task based on the example cases above.
[Task]
{Here is the task description of the current task}
[train.py] ```python
{Here is the original Python script of the current task}
```
Start the python code with "```python". Please ensure the completeness of the code so that it can be run without additional modifications.  
\end{lstlisting}
}
\end{tcolorbox}


\end{document}